%% file: spatial_evolutionary_GAN.tex
\documentclass[sigconf]{acmart}

\usepackage{matharticle}
\usepackage{algpseudocode}

\usepackage{paralist}
\usepackage{todonotes}
\usepackage{xspace}
\usepackage{booktabs} 
\usepackage[utf8]{inputenc} 
\usepackage[T1]{fontenc}    
\usepackage{hyperref}       
\usepackage{url}            
\usepackage{amsfonts}       
\usepackage{nicefrac}       
\usepackage{microtype}      
\usepackage{comment}

\algblock{ParFor}{EndParFor}
\algnewcommand\algorithmicparfor{\textbf{parfor}}
\algnewcommand\algorithmicpardo{\textbf{do}}
\algnewcommand\algorithmicendparfor{\textbf{end\ parfor}}
\algrenewtext{ParFor}[1]{\algorithmicparfor\ #1\ \algorithmicpardo}
\algrenewtext{EndParFor}{\algorithmicendparfor}

\newcommand{\ec}{evolutionary computation\xspace}

\newcommand{\EGAN}{\textbf{E-GAN}\xspace}
\newcommand{\DCGAN}{\textbf{GAN-BCE}\xspace}
\newcommand{\SEGAN}{\textbf{Mustangs}\xspace}
\newcommand{\SCoevGANmm}{\textbf{Lip-BCE}\xspace}
\newcommand{\SCoevGANls}{\textbf{Lip-MSE}\xspace}
\newcommand{\SCoevGANh}{\textbf{Lip-HEU}\xspace}
\newcommand{\Lipizzaner}{\textbf{Lipizzaner}\xspace}
\newcommand{\Lipi}{\textbf{Lipizzaner}\xspace}

\newcommand{\mutationDiversity}{mutation diversity\xspace}
\newcommand{\SEGANlong}{\emph{MUtation SpaTial gANs}\xspace}

\usepackage{mwe}
\usepackage{graphicx}

\setcopyright{none}

\acmConference[GECCO '19]{Genetic and Evolutionary Computation Conference}{July 13--17, 2019}{Prague, Czech Republic}
\acmBooktitle{Genetic and Evolutionary Computation Conference (GECCO '19), July 13--17, 2019, Prague, Czech Republic}
\acmDOI{10.1145/3321707.3321860}

\begin{document}
\title{Spatial Evolutionary Generative Adversarial Networks}

\author{Jamal Toutouh}
\affiliation{%
  \institution{Massachusetts Institute of Technology, CSAIL}
  \streetaddress{Cambridge, MA}
}
\email{toutouh@mit.edu}

\author{Erik Hemberg}
\affiliation{%
  \institution{Massachusetts Institute of Technology, CSAIL}
  \streetaddress{Cambridge, MA}
}
\email{hembergerik@csail.mit.edu}

\author{Una-May O'Reilly}
\affiliation{%
  \institution{Massachusetts Institute of Technology, CSAIL}
  \streetaddress{Cambridge, MA}
}
\email{unamay@csail.mit.edu}

\renewcommand{\shortauthors}{Toutouh et al.}

\input{abstract}

%
%



\maketitle

\input{introduction}
\input{related-work}

\input{system}

\input{experiments}
\input{results}

\input{conclusions}
\input{acks}

\bibliographystyle{ACM-Reference-Format}
\bibliography{spatial_evolutionary_GAN.bib} 

\end{document}

%% file: abstract.tex
\begin{abstract}
Generative adversary networks (GANs) suffer from training pathologies such as instability and mode collapse. These pathologies mainly arise from a lack of diversity in their adversarial interactions. Evolutionary generative adversarial networks apply the principles of evolutionary computation to mitigate these problems. We hybridize two of these approaches that promote training diversity. One, \EGAN, at each batch, injects mutation diversity by  training the (replicated) generator with three independent objective functions then selecting the resulting best performing generator for the next batch. The other, \Lipi, injects population diversity by training a two-dimensional grid of GANs with a distributed evolutionary algorithm that includes neighbor exchanges of  additional training adversaries, performance based selection and population-based hyper-parameter tuning. We propose to combine mutation and population approaches to diversity improvement.  We contribute a superior evolutionary GANs training method, \SEGAN, that eliminates the single loss function used across \Lipi’s grid.  Instead, each training round, a loss function is selected with equal probability, from among the three \EGAN uses. Experimental analyses on standard benchmarks, MNIST and CelebA, demonstrate that \SEGAN provides a statistically faster training method resulting in more accurate networks.
\end{abstract}

%% file: introduction.tex
\section{Introduction}
\label{sec:introduction}

Generative adversarial networks (GANs) have emerged as a powerful
machine learning paradigm. They were first introduced for the task of estimating
a distribution function underlying a given set of samples~\cite{goodfellow2014generative}. 
A GAN consists of two neural networks, one a generator and the other a
discriminator. The discriminator is trained to correctly discern the
``natural/real'' samples from ``artificial/fake'' samples produced by
the generator. The generator, given a latent random space, is trained
to transform its inputs into samples that fool the
discriminator. Formulated as a minmax optimization problem through the
definitions of discriminator and generator loss, training can converge
on an optimal generator, one that approximates the latent true
distribution so well that the discriminator can only provide a label at random for any sample.  

The early successes of GANs in generating realistic, complex, multivariate distributions
motivated a growing body of applications, such as image
generation~\cite{gan2017triangle}, video
prediction~\cite{liang2017dual}, image
in-painting~\cite{yeh2017semantic}, and text to image
synthesis~\cite{reed2016learning}. 
However, while the competitive juxtaposition of the generator and discriminator is a compelling design, 
GANs are notoriously hard to train. 
Frequently training dynamics show pathologies. Since the generator and the discriminator are differentiable networks, optimizing the minmax GAN objective is
generally performed by (variants of) simultaneous gradient-based
updates to their parameters~\cite{goodfellow2014generative}.  This
type of gradient-based training rarely converges to an equilibrium.
GAN training thus exhibits degenerate behaviors, such as \emph{mode
collapse}~\cite{arora2017gans}, \emph{discriminator
collapse}~\cite{li2017towards}, and \emph{vanishing
gradients}~\cite{arjovsky2017towards}.

Different objectives impact the gradient information used to update parameters weights of the networks. 
Therefore, changing the objective impacts the search trajectory and could eliminate or decrease the frequency of pathological trajectories. A set of recent studies by members of the machine learning community proposed different objective functions.  Generally, these
functions compute loss as the distance between the fake data and real data
distributions according to different measures. 
 The original GAN~\cite{goodfellow2014generative} applies the \emph{Jensen-Shannon divergence (JSD)}.  Other measures include:
\begin{inparaenum} [\itshape 1)]
\item \emph{Kullback-Leibler divergence (KLD)}~\cite{nguyen2017dual}, 
\item the \emph{Wasserstein distance}~\citep{arjovsky2017wasserstein}, 
\item the \emph{least-squares (LS)}~\citep{mao2017least}, and
\item the \emph{absolute deviation}~\cite{zhao2016energy}.
\end{inparaenum}

Each of these objective functions improves training but none entirely solves all of its challenges. An evolutionary computation project investigated an evolutionary generative adversarial network (\EGAN), a different approach~\cite{wang2018evolutionary}. \EGAN, batch after batch, is able to guide its trajectory with gradient information from a population of three different objectives, which defines the gradient-based mutation to be applied. 
As we will describe in more detail in Section~\ref{sec:related-work}, each batch, \EGAN trains each of three copies of the GAN with one of the three objectives in the population. 
After this independent training, \EGAN selects the best GAN according to a given fitness function to start the next batch and train further. This process splices batch-length trajectories from different gradient information together. Essentially, \EGAN injects \textit{\mutationDiversity} into training. 
As a result, on some benchmarks, \EGAN improves and provides comparable performance on a baseline using a single objective. 


%
 
Another \ec idea for addressing training pathologies comes from competitive coevolutionary algorithms. 
With one population adversarially posed against another, they optimize with a minmax objective like GANs. 
Pathologies similar to what is reported in GAN training have been observed in coevolutionary algorithms, such as \emph{focusing}, \emph{relativism}, and \emph{lost of gradient}~\cite{popovici2012coevolutionary}. 
They have been attributed to a lack of diversity. Each population converges or the coupled population dynamics lock into a tit-for-tat pattern of ineffective signaling.  Spatial distributed populations have been demonstrated to be particularly effective at resolving this. This approach has been transferred to GANs with 
\Lipizzaner~\cite{schmiedlechner2018lipizzaner,schmiedlechner2018towards}. \Lipi uses a spatial distributed competitive coevolutionary algorithm.  It places the individuals of the generator and discriminator populations on a grid (each cell contains a pair of generator-discriminators). 
Each generator is evaluated against all the discriminators of its neighborhood and 
the same happen with each discriminator. \Lipizzaner takes advantage of neighborhood communication to propagate models. 
 \Lipizzaner, in effect, provides diversity in the \emph{genome space}.  

In this paper we ask whether a method that capitalizes on ideas from both \EGAN and \Lipizzaner is better than either one of them. Specifically, can a combination of diversity in mutation and genome space train GANs  faster, more accurately and more reliably?
Thus, we present the \SEGANlong training method, \SEGAN. 
For each cell of the grid, \SEGAN selects randomly with equal probability a given loss function from among the set of three that \EGAN introduced, which is applied for the current training batch. 
This process is repeated for each batch during the whole GAN training. 
We experimentally evaluate \SEGAN on standard benchmarks, MNIST and CelebA, to determine whether it 
provides more accurate results and/or requires shorter execution times. The main contributions of this paper are:
\begin{inparaenum}[\itshape 1)]
\item \SEGAN, a training method of GANs that provides both mutation and genome diversity.
\item A open source software implementation of \SEGAN\footnote{\SEGAN source code - \texttt{https://github.com/mustang-gan/mustang}}, 
\item A demonstration of \SEGAN's higher accuracy and performance on MNIST and CelebA.
\item A deployment of \SEGAN on cloud computing infrastructure that optimizes the GAN grid in parallel. 
\end{inparaenum}

The rest of the paper is organized as
follows. Section~\ref{sec:related-work} presents related
work. Section~\ref{sec:system} describes the method. The experimental
setup and the results are in sections~\ref{sec:experiments}
and~\ref{sec:results}. Finally, conclusions are drawn and future work
is outlined in Section~\ref{sec:conclusions}.

%% file: related-work.tex
\section{Related work}
\label{sec:related-work}

Recent work have focused on improving the robustness of GAN training and the overall quality of generated samples~\cite{arjovsky2017towards,arora2017gans,nguyen2017dual}. 
Prior approaches tried to mitigate degenerate GAN dynamics by using heuristics, 
such as decreasing the optimizer's learning rate along the iterations~\cite{Radford2015unsupervised}. 
Other authors have proposed changing generator's or discriminator's objectives~\cite{arjovsky2017wasserstein,mao2017least,nguyen2017dual,zhao2016energy}. 
More advanced methods apply ensemble approaches~\cite{wang2016ensembles}. 

A different category of studies employ multiple generators and/or discriminators.   
Some remarkable examples analyze training a cascade of GANs~\cite{wang2016ensembles}; 
sequentially training and adding new generators with boosting techniques~\cite{tolstikhin2017adagan}; training in parallel multiple generators and discriminators~\cite{jiwoong2016generative}; and 
training an array of discriminators specialized in a different low-dimensional projection of the data~\cite{neyshabur2017stabilizing}. 

Recent work by Yao and co-authors proposed \EGAN, whose main idea is to evolve a population of three independent loss functions defined according to three distance metrics (JSD, LS, and a metric based on JSD and KL)~\cite{wang2018evolutionary}. 
One at a time, independently, the loss functions are used to train a generator from some starting condition, over a batch. 
The generators produced by the loss functions are evaluated by a single discriminator (considered optimal) that returns a fitness value for each generator.
The best generator of the three options is then selected and training continues, with the next training batch, and the three different loss functions. 
The use of different objective functions (mutations) overcomes the limitations of a single individual training objective and a better adapts the population to the evolution of the discriminator. 
\EGAN defines a specific fitness function that evaluates the generators in terms of the quality and the diversity of the generated samples.
The results shown that \EGAN is able to get higher inception scores, while showing comparable stability when it goes to convergence. \EGAN works because the evolutionary population injects diversity into the training. Over one training run, different loss functions inform the best generator of a batch.

Another evolutionary way to improve training is also motivated by diversity~\cite{schmiedlechner2018lipizzaner,schmiedlechner2018towards}.
\Lipizzaner simultaneously trains a spatially distributed population of GANs (pairs of generators and discriminators) that allows neighbors to communicate and share information. 
Gradient learning is used for GAN training and evolutionary selection and variation is used for hyperparameter learning.
%
Overlapping neighborhoods and local communication allow efficient propagation of improving models. 
Besides, this strategy has the ability to distribute the training process on parallel computation architectures, and therefore, it can efficiently scale.   
\Lipizzaner coevolutionary dynamics are able to escape degenerate GAN training behaviors, 
e.g, mode collapse and vanishing gradient, and resulting generators provide accurate and diverse samples. 

In this paper we ask whether an advance that capitalizes on ideas from both \EGAN (i.e., diversity in the mutation space) and \Lipizzaner (i.e., diversity in genomes space) is better than either one of them. 


%% file: system.tex
\begin{figure*}[!h]
\includegraphics[width=\linewidth]{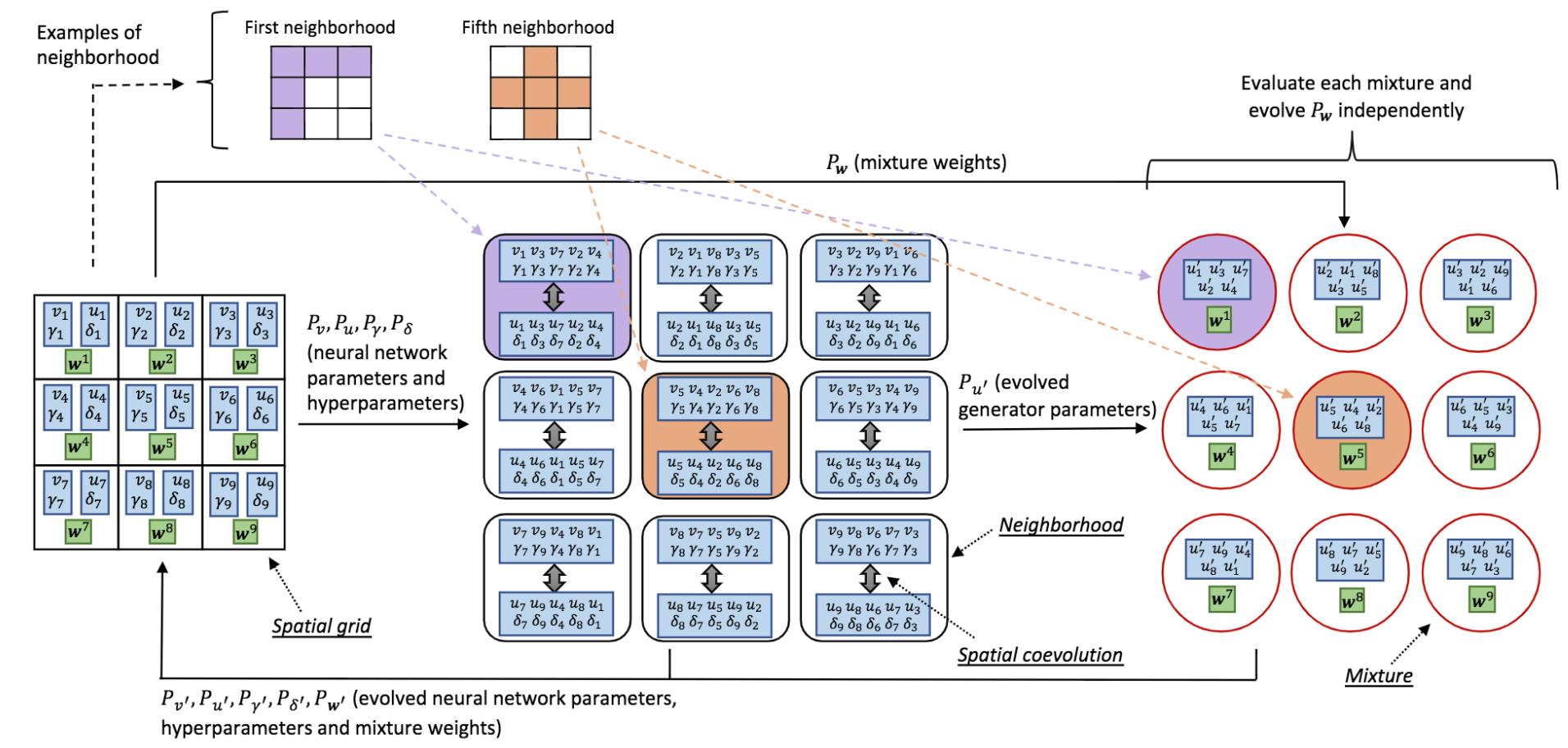}
\caption{\small Spatial coevolution training on a
		$3 \times 3$ grid~\cite{schmiedlechner2018towards}. $P_v = \{v_1, \dots v_9\}$ and $P_u = \{u_1, \dots
		u_9\}$ denote neural network weights of discriminator and
		generator population respectively. $P_{\gamma} = \{\gamma_1,
		\dots, \gamma_9\}$ and $P_{\delta} = \{\delta_1, \dots, \delta_9\}$
		denote the hyperparameters (e.g., learning rate) of discriminator
		and generator population, respectively. $P_{\bf w} = \{\mathbf{w}_1, \dots, \mathbf{w}_9\}$
		denote the mixture weights. The $(\cdot)'$ notation denotes the value of $(\cdot)$ after one iteration of coevolution.}
\label{fig:lipizzaner}
\end{figure*}

\section{\SEGAN Method}
\label{sec:system}

This section presents \SEGAN devised in this work. 
First, we introduce the general optimization problem of GAN training. Then, we describe a method for spatial coevolution GANs training. Finally, we present the multiple mutations applied to produce the generators offspring.

\subsection{General GAN Training}

In this paper we adopt a mix of notation used in~\cite{li2017towards}. 
Let $\calG=\{G_u, u \in \calU\}$ and $\calD=\{D_v, v \in \calV\}$ denote the class of generators and discriminators, respectively, where $G_u$ and $D_v$ are functions parameterized by $u$ and $v$. 
$\calU, \calV \subseteq \R^{p}$ represent the parameters space of the generators and discriminators, respectively.  Further, let $G_*$ be the target unknown distribution that we would like to fit our generative model to. 

Formally, the goal of GAN training is to find parameters $u$ and $v$ so as to optimize the objective function
\begin{equation}
\label{eq:gan-def}
\min_{hu\in \calU}\max_{v \in \calV} \calL(u,v)\;, \;\text{where} \nonumber
\end{equation}
\begin{equation}
\calL(u,v) = \E_{x\sim G_*}[\phi (D_v(x))] + \E_{x\sim G_u}[\phi(1-D_v(x))]\;,
\end{equation}
and $\phi:[0,1] \to \R$, is a concave function, commonly referred to as the \emph{measuring function}. 
In practice, we have access to a finite number of training samples $x_1, \ldots, x_m \sim G_*$. 
Therefore,
an empirical version $\frac{1}{m}\sum_{i=1}^{m} \phi(D_v(x_i))$ is used to estimate $\E_{x\sim G_*}[\phi (D_v(x))]$. The same holds for $G_u$.

\subsection{Spatial Coevolution for GAN Training}
\label{sec:coev-training}

Evolutionary computation implements mechanisms inspired by biological evolution such as reproduction, diversity generation, and survival of the fittest to address optimization problems. 
In this case. we apply the competitive coevolutionary algorithm outlined in Algorithm~\ref{alg:basic-coev-gan} to optimize GANs. It evolves two populations, $P_u=\{u_1, \ldots, u_T\}$ a population of generators and $P_v=\{v_1, \ldots, v_T\}$ a population of discriminators to create diversity in genomes spaces, where $T$ is the population size. 
The fitness $\calL$ of each generator $u_i\in P_u$ and discriminator $v_j \in P_v$ are assessed according to their interactions with a set of discriminators from $P_v$ and  generators from $P_u$, respectively (Lines~\ref{line:evl-beg} to~\ref{line:evl-end}). 
The fittest individuals are used to generate the new of individuals (generators and discriminators) by applying mutation (see Section~\ref{sec:segan-mutation}). 
The new individuals replace the ones in the current population if they perform better (better fitness) to produce the next generation.  

\SEGAN applies the spatially distributed coevolution summarized in Algorithm~\ref{alg:basic-coev-gan}, the individuals of both populations (generators of $P_u$ and discriminators of $P_v$) are distributed on the cells of a two imensional toroidal grid~\cite{schmiedlechner2018towards}. 
Spatial coevolution has shown a considerable ability in maintaining diversity in the populations and fostering continuous arms races~\cite{Mitchell06therole,Williams2005}.  

\begin{algorithm}[h!]
	\floatname{algorithm}{\footnotesize Algorithm}
	\footnotesize
	\caption{\footnotesize  \texttt{GeneralCoevGANs}($P_u, P_v, \calL, \{\alpha_{i}\}, \{\beta_{i}\}, p_{\mu}, $) \newline
		\textbf{Input:} \newline  
		~$P_u$~: generator population \hspace{11em} 
		~$P_v$~: discriminator population \newline 
		~$\{\alpha_{i}\}$~: selection probability\hspace{10em}
		~$\{\beta_{i}\}$~: mutation probability \newline 
		~$I$~: number of generations \hspace{11em}
		~$\calL$~: GAN objective function \newline 
		~$p_{\mu}$~: objective-selection probabilities vector \newline 
		\textbf{Return:} \newline 
		~$P_u$~: evolved generator population \newline 
		~$P_v$~: evolved discriminator population \newline 
	}
	\label{alg:basic-coev-gan}
	\begin{algorithmic}[1]
		 \bf
		\For {$i$ in $[1 \ldots I]$}
		\Statex \hspace{1.5em}// Evaluate $P_u$ and $P_v$
		\State $f_{u_1\ldots u_T} \gets 0$ \label{line:evl-beg}
		\State $f_{v_1\ldots v_T} \gets 0$
		\For {each $u_i$ in $P_u$, each $v_j$ in $P_v$}
		\State $f_{u_i} \mathrel{-}= \calL(u_i, v_j)$ \label{line:fit-update-beg}
		\State $f_{v_j} \mathrel{+}= \calL(u_i, v_j)$ \label{line:fit-update-end}
		\EndFor  \label{line:evl-end}
		\Statex \hspace{1.5em}// Sort $P_u$ and $P_v$
		\State $u_{1\ldots T} \gets u_{s(1)\ldots s(T)} \texttt{ with } s(i)=\texttt{argsort} (f_{u_1\ldots u_T} , i)$ \label{line:sel-beg}
		\State $v_{1\ldots T} \gets v_{s(1)\ldots s(T)} \texttt{ with } s(j)=\texttt{argsort} (f_{v_1\ldots v_T} , j)$
		\Statex \hspace{1.5em}// Selection 
		\State $u_{1\ldots T} \gets u_{s(1)\ldots s(T)}  \texttt{ with } s(i)=\texttt{argselect} (u_{1\ldots T}  , i, \{\alpha_i\})$
		\State $v_{1\ldots T}  \gets v_{s(1)\ldots s(T)} \texttt{ with } s(j)=\texttt{argselect} (v_{1\ldots T}, j, \{\alpha_{j}\})$ \label{line:sel-end}
		\Statex \hspace{1.5em}// Mutation \& Replacement
		\State $u_{1\ldots T}\gets \texttt{replace}(\{u_{i}\},\{u^\prime_{i}\}) \texttt{ with } u^\prime_i=\texttt{mutateP$_u$} (u_i, \{\beta_{i}\})$ \label{line:mut-beg}
		\State $v_{1\ldots T}  \gets \texttt{replace}(\{v_{j}\},\{v^\prime_{j}\}) \texttt{ with } v^\prime_j=\texttt{mutateP$_v$} (v_j, \{\beta_{i}\})$ \label{line:mut-end}
		\EndFor
		\State \Return $P_u, P_v$ \label{line:basic-coev-return}
	\end{algorithmic}

\end{algorithm}

The cell's \emph{neighborhood} defines the subset of individuals of $P_u$ and $P_v$ to interact with and it is specified by its size $s_{n}$.
Given a $m\times n$-grid, there are $m\times n$ neighborhoods. 
Without losing generality, we consider $m^2$ square grids to simplify the notation.  
In our study, we use a five-cell neighborhood, i.e, one center and four adjacent cells (see Figure~ \ref{fig:lipizzaner}). 
We apply the same notation used in~\cite{schmiedlechner2018towards}. 
For the $k$-th neighborhood in the grid, we refer to the generator in its center cell by $P^{k,1}_u\subset P_u$ and the set of generators in the rest of the neighborhood cells by $P^{k,2}_u, \ldots, P^{k,s_n}$, respectively. Furthermore, we denote the union of these sets by $P^k_u = \cup^{s_n}_{i=1} P^{k,i}_u \subseteq P_u$, which represents the $k$th generator neighborhood.

In the spatial coevolution applied  here, each neighborhood performs an instance of Algorithm~\ref{alg:basic-coev-gan} with the populations $P^k_u$ and $P^k_v$ to update its center cell, i.e. $P^{k,1}_u$, $P^{k,1}_v$, with the returned values~(Line~\ref{line:basic-coev-return} of Algorithm~\ref{alg:basic-coev-gan}). 

Given the $m^2$ neighborhoods, all the individuals of $P_u$ and $P_v$ will get updates as  $P_u = \cup^{m^2}_{k=1}P^{k}_u$, $P_v=\cup^{m^2}_{k=1}P^{k}_v$. 
These $m^2$ instances of Algorithm~\ref{alg:basic-coev-gan} run in parallel in an asynchronous fashion when dealing with reading/writing from/to the populations. 
This implementation scales with lower communication overhead, allows the cells to run its instances without waiting each other, increases the diversity by mixing individuals computed during different stages of the training process, and performs better with limited number of function evaluations~\cite{nielsen2012novel}. 

Taking advantage of the population of $|P_u|$ generators trained, 
the spatial coevolution method selects one of the generator neighborhoods 
$\{P^k_u\}_{1\leq k \leq m^2}$ as a mixture of generators according to a given performance metric $f:\calU^{s_n} \times \R^{s_n}  \to \R$.   
Thus, it is chosen the best generator mixture $P^*_u\in \calU^{s_n} $ according to the mixture weights $\vw^* \in [0,1]^{s_n}$. 
Hence, the $s_n$-dimensional mixture weight vector $\vw$ is defined as follows
\begin{equation}
\footnotesize
P^*_u, \vw^* =\argmax_{P^{k}_u, \vw^k: 1\leq k \leq m^2  } f\big(\sum_{{u_i \in P^{k}_u \\ w_i \in \vw^k}} w_i G_{u_i}\big)\;,
\label{eq:mixture}
\end{equation}
where $w_i$ represents the mixture weight of (or the probability that a data point comes from) the $i$th generator in the neighborhood, with $\sum_{w_i \in \vw^k} w_i = 1$. 
These hyperparameters $\{\vw^k\}_{1\leq k \leq m^2}$ are optimized during the training process after each step of spatial coevolution by applying an evolution strategy (1+1)-ES~\cite{schmiedlechner2018towards}.


\subsection{\SEGAN Gradient-based Mutation}
\label{sec:segan-mutation}

\SEGAN coevolutionary algorithm generates the offspring of both populations $P_h$ and $P_q$ by by applying \emph{asexual reproduction}, i.e. next generation's of individuals are produced by applying mutation (Lines~\ref{line:mut-beg} and~\ref{line:mut-end} in Algorithm~\ref{alg:basic-coev-gan}). 
These mutation operators are defined according to a giving training objective, which generally attempts to minimize the distance between the generated fake data and real data distributions according to a given measure. 
\Lipizzaner applies the same gradient-based mutation for both populations during the coevolutionary learning~\cite{schmiedlechner2018towards}. 

In this study, we add mutation diversity to the genome diversity provided by \Lipizzaner. 
Thus, we use the mutations used by \EGAN to generate the offspring of generators~\cite{wang2018evolutionary}. 
E-GAN applies three different mutations corresponding with three different minimization objectives w.r.t. the generator:
\begin{inparaenum}[\itshape 1)]
	\item \emph{Minmax mutation}, which objective is to minimize the JSD between the real and fake data distributions, i.e., $JSD(p_{real}\parallel p_{fake})$ (see Equation~(\ref{eq:minmax-mut})).  
	\item \emph{Least-square mutation}, which is inspired in the least-square GAN~\cite{mao2017least} that applies this criterion to adapt both, the generator and the discriminator. The objective function is formulated as shown in Equation~(\ref{eq:leastsquare-mut}).  
	\item \emph{Heuristc mutation}, which maximizes the probability of the discriminator being mistaken by minimizing the objective function in Equation~(\ref{eq:heuristic-mut}). This objective is equal to minimizing $[KL(p_{real}\parallel p_{fake}) - 2JSD(p_{real}\parallel p_{fake})]$  
\end{inparaenum}

\begin{equation}
\label{eq:minmax-mut}
M^{minmax}_G = \frac{1}{2} \E_{x\sim G_u}[log(1-D_v(x))]
\end{equation}

\begin{equation}
\label{eq:leastsquare-mut}
M^{least-square}_G = \E_{x\sim G_u}[log(D_v(x) - 1)^2]
\end{equation}

\begin{equation}
\label{eq:heuristic-mut}
M^{heuristic}_G = \frac{1}{2} \E_{x\sim G_u}[log(D_v(x))]
\end{equation}

Thus, the mutation applied to the generators (\texttt{mutateP$_u$}) in Line~\ref{line:mut-beg} of Algorithm~\ref{alg:basic-coev-gan} is defined in the Algorinthm~\ref{alg:alg:segan-mutation}. 
The new generator is produced by using a loss function (mutation) to optimize one of the three objectives functions introduced above. i.e., $M^{minmax}_G$, $M^{least-square}_G$, and $M^{heuristic}_G$, which are \emph{binary cross entropy (BCE) loss}, \emph{mean square error (MSE) loss}, and \emph{heuristic} loss, respectively.

\begin{figure}[!h]
\includegraphics[width=\linewidth, trim={2cm 5.2cm 5cm 1.6cm},clip]{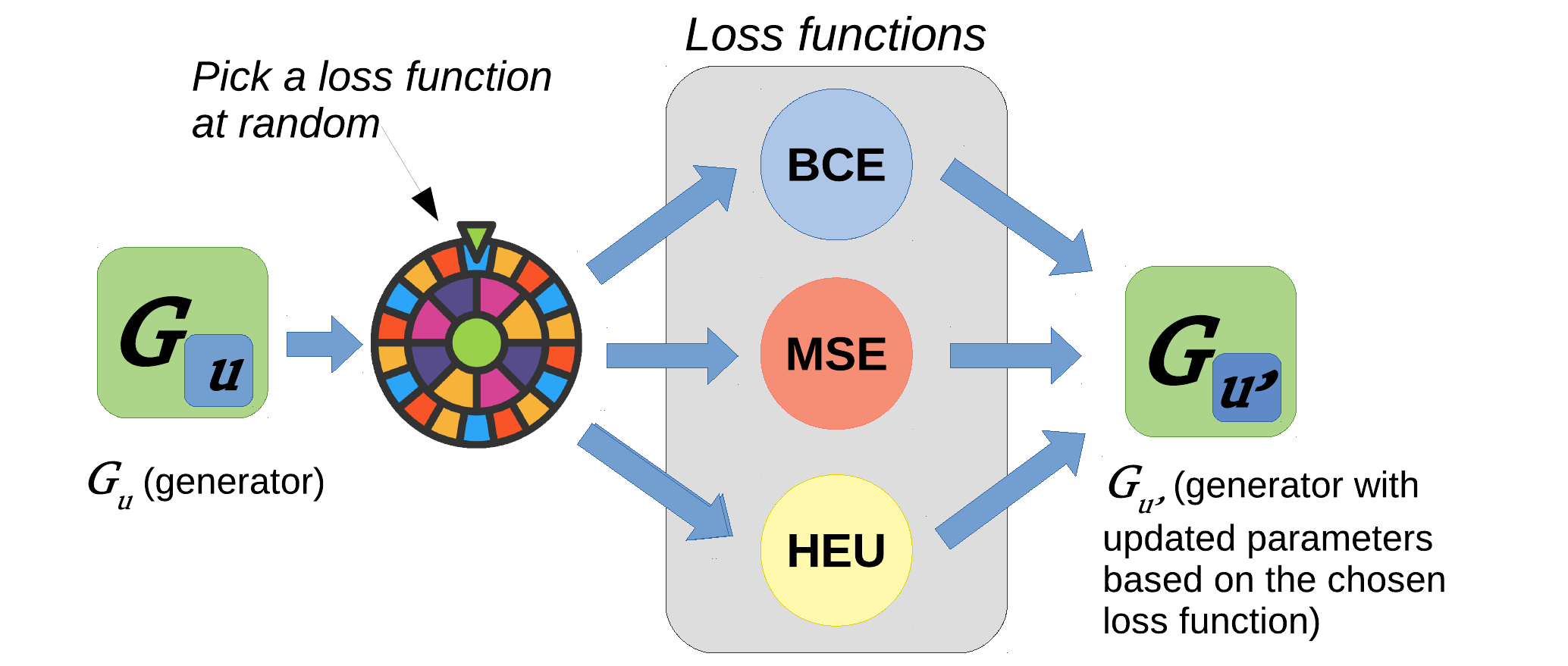}
\caption{\small Graphical representation of the mutation used in \SEGAN. The generator (ancestor) $G_u$ is mutated to produce the new generator $G_{u\prime}$ by using one of the loss functions chosen at random.}
\label{fig:segan-mutation}
\end{figure}

\EGAN applies the three mutations to the generator (ancestor) and it selects the individual that provides the best fitness when is evaluated against the discriminator~\cite{wang2018evolutionary}. 
In contrast, \SEGAN picks at random with same the probability ($\frac{1}{3}$) one of the mutations (loss functions), as it is shown in Figure~\ref{fig:segan-mutation}, and then the gradient descent method is applied accordingly (see Algorithm~\ref{alg:alg:segan-mutation}). 
This enables diversity in the mutation space without adding noticeable overhead over the spatial coevolutionary training method presented before, 
since \SEGAN evaluates only the mutated generator instead of three as \EGAN does. 

\begin{algorithm}[h!]
	\floatname{algorithm}{\footnotesize Algorithm}
	\footnotesize
	\caption{\footnotesize  \texttt{mutateP$_u$}($u$) \newline
		\textbf{Input:} \newline  
		~$u$~: generator parameters\newline  
		\textbf{Return:} \newline 
		~$u^\prime$~: mutated generator parameters
	}
	\label{alg:alg:segan-mutation}
	\begin{algorithmic}[1]
		 \bf
		 \State $operator \gets \texttt{pick\_random}[1 \ldots 3]$ \label{line:mutation-picking}
      	\If {$operator == 1$}
		\State $u^\prime \gets \texttt{applyGradienDescent}(M^{minmax}_G, u)$
		\ElsIf {$operator == 2$}
		\State $u^\prime \gets \texttt{applyGradienDescent}(M^{least-square}_G, u)$
		\Else
		\State $u^\prime \gets \texttt{applyGradienDescent}(M^{heuristic}_G, u)$
		\EndIf
		\State \Return $u^\prime$ \label{line:mutation-return}
	\end{algorithmic}
\end{algorithm}

\EGAN, \Lipizzaner, and \SEGAN apply the same mutation (loss function) to update the discriminators, the one defined to address the GAN minmax optimization problem described in Equation~(\ref{eq:gan-def}).

%% file: experiments.tex
\vspace{-0.2cm}
\section{Experimental Setup}
\label{sec:experiments}

\SEGAN is evaluated on two common image data sets: MNIST\footnote{The MNIST Database  - \texttt{http://yann.lecun.com/exdb/mnist/}} and CelebA\footnote{The CelebA Database  - \texttt{http://mmlab.ie.cuhk.edu.hk/projects/CelebA.html}}. 
The experiments compare the following methods:
\begin{asparaenum}
\item [\DCGAN] a standard GAN which uses $M^{minmax}_G$ objective 
\item [\EGAN] the \EGAN method
\item [\SCoevGANmm] \Lipizzaner with $M^{minmax}_G$ objective
\item [\SCoevGANls] \Lipizzaner with $M^{least-square}_G$ objective
\item [\SCoevGANh] \Lipizzaner with $M^{heuristic}_G$ objective
\item [\SEGAN] the \SEGAN method
\end{asparaenum}

The settings used for the experiments are summarized in Table~\ref{tab:net-arq}.   
The four spatial coveolutionary GANs use
the parameters presented in Table~\ref{tab:net-arq}.
\EGAN and \DCGAN both use the Adam optimizer with an initial learning rate (0.0002). The other parameters of \EGAN use the same configuration as used in~\citep{wang2018evolutionary}. 

For MNIST data set experiments, all methods use the same stop condition: a computational budget of nine hours (9h).  The distributed methods
of \SEGAN, \SCoevGANmm, \SCoevGANls, and \SCoevGANh use a grid size of
$3\times3$, and are able to train nine networks in parallel. Thus,
they are executed during one hour to comply with the computational
budget of nine hours. 
Regarding CelebA experiments, the four spatial coveolutionary GANs are analyzed. Thus, they stop after performing 20~training epochs, since they require similar computational budget to perform them. 

All methods have been implemented in \texttt{Python3}
and \texttt{pytorch}\footnote{Pytorch Website
- \texttt{https://pytorch.es/}}. The spatial coevolutionary ones have
extended the \Lipizzaner
framework~\cite{schmiedlechner2018lipizzaner}.

The experimental analysis is performed on a cloud computation platform
that provides 8~Intel Xeon cores 2.2GHz with 32 GB RAM and a NVIDIA Tesla T4 GPU with 16 GB RAM. 
We run multiple independent runs for each method. 
   
\begin{table}[!h]
	\centering
	\small
	\caption{\small Network topology of the GANs trained.}
	\label{tab:net-arq}
	\begin{tabular}{lrr}
	    \toprule
		\textbf{Parameter} & \textbf{MNIST} & \textbf{CelebA}  \\ \hline
		\multicolumn{3}{c}{\textit{Network topology}} \\ \hline
		Network type  & MLP & DCGAN \\ 
		Input neurons & 64 & 100 \\ 
		Number of hidden layers & 2 & 4  \\ 
		Neurons per hidden layer & 256 & 16,384 - 131,072  \\ 
		Output neurons  & 784 & 64$\times$64$\times$64\\ 
		Activation function & $tanh$ & $tanh$ \\ \hline
		\multicolumn{3}{c}{\textit{Training settings}} \\ \hline
		Batch size  & 100  & 128 \\ 
		Skip N disc. steps & 1 & -\\ \hline
		\multicolumn{3}{c}{\textit{Coevolutionary settings}} \\ \hline
		Stop condition & 9 hours comp. & 20 training epochs \\ 
		Population size per cell & 1  & 1 \\ 
		Tournament size & 2  & 2\\ 
		Grid size & 3$\times$3  & 2$\times$2  \\ 
		Performance metric ($m$) & FID & FID \\
		Mixture mutation scale & 0.01 & 0.01 \\ \hline
		\multicolumn{3}{c}{\textit{Hyperparameter mutation}} \\ \hline
		Optimizer & Adam  & Adam \\ 
		Initial learning rate & 0.0002  & 0.00005 \\ 
		Mutation rate & 0.0001 & 0.0001  \\ 
		Mutation probability & 0.5 & 0.5 \\ 
		\bottomrule
	\end{tabular}
\end{table}

For quantitative assessment of the accuracy of the generated fake data the Frechet inception distance (FID) is evaluated~\cite{heusel2017gans}. 
We analyze the computational performance of each method.  
Finally, we evaluate the diversity of the data samples generated. 

%% file: results.tex
\section{Results and Discussion}
\label{sec:results}

This section presents the results and the analyses of the studied optimization methods. 
The first three subsections evaluate the MNIST experiments in terms of the FID score, the computational performance, and the diversity of the generated samples, respectively. 
The last one analyzes the CelebA results in terms of FID score.  

\subsection{Quality of the Generated Data}
\label{sec:fid}

\begin{figure}[!h]
\includegraphics[width=\linewidth]{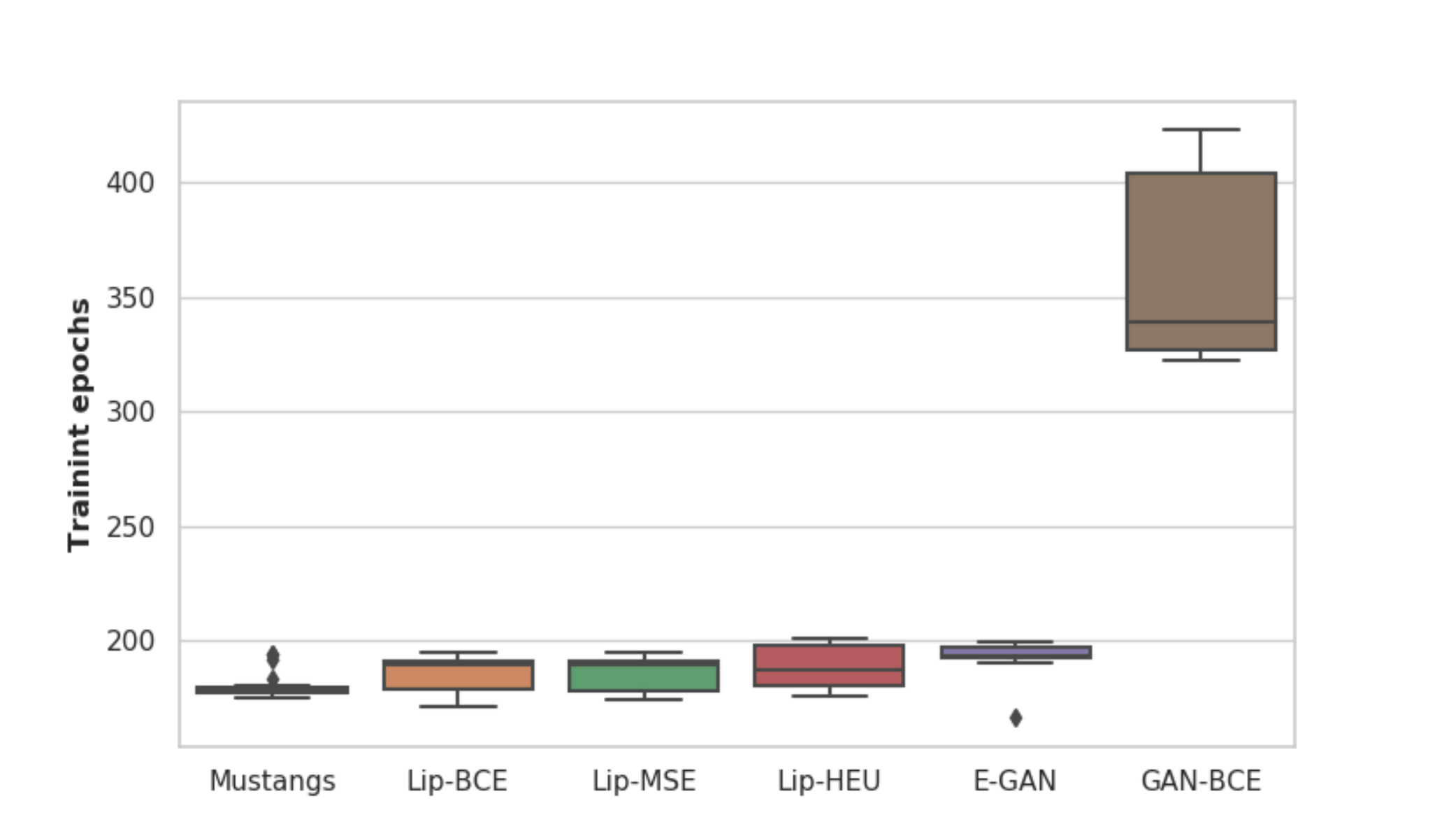}
\vspace{-0.1cm}
\caption{Results on MNIST dataset. Boxplot that shows the best FIDs computed for each independent run. }
\label{fig:fid-boxplot}
\vspace{-0.2cm}
\end{figure}

Table~\ref{tab:fid-results} shows the best FID value from each of the 30 independent runs performed for each method. \SEGAN has the lowest median (see Figure~\ref{fig:fid-boxplot} for a boxplot). All the methods that used
\Lipizzaner are better than \EGAN and \DCGAN. However, there is quite
a significant increase in FID value between the \SCoevGANls and the
other \Lipizzaner based methods. The results indicate that \SEGAN is
robust to the varying performance of the individual loss functions and
can still find a high performing mixture of generators. This helps to
strengthen the idea that diversity, both in genome and mutation space,
provides robust GAN training. 

\begin{table}[!h]
	\centering
	\small
	\caption{\small FID MNIST results in terms of best mean, normalized standard deviation, median and interquartile range (IQR). (Low FID indicates good performance)}
	\vspace{-0.2cm}
	\label{tab:fid-results}
	\begin{tabular}{lrrrr}
	    \toprule
		\textbf{Algorithm} & \textbf{Mean} & \textbf{Std\%} & \textbf{Median} & \textbf{IQR}  \\ \hline
\SEGAN & \textit{42.235} & 12.863\% & \textit{43.181} &  7.586 \\ 
\SCoevGANmm & 48.958 & 20.080\% & 46.068 &  4.663 \\ 
\SCoevGANls & 371.603 & 20.108\% & 381.768 &  104.625 \\ 
\SCoevGANh & 52.525 & 17.230\% & 52.732 &  9.767 \\ 
\EGAN & 466.111 & 10.312\% & 481.610 &  69.329 \\ 
\DCGAN & 457.723 & 2.648\% & 459.629 &  17.865 \\ 
		\bottomrule
	\end{tabular}
	\vspace{-.1cm}
\end{table}

The results provided by the methods that generates diversity in genome
space only (\SCoevGANmm, \SCoevGANh, and \SCoevGANls) are significantly
more competitive than \EGAN, which provides diversity in mutation
space only (see Figure~\ref{fig:fid-boxplot}). 
Therefore, the spatial distributed coevolution provides an
efficient tool to optimize GANs.

Surprisingly the median FID score of \DCGAN is better than
\EGAN. \EGAN has a larger variance of FID scores compared to \DCGAN,
and in the original paper it was shown that \EGAN performance improved
with more epochs (by using a computational budget of 30h, compared to
the 9h we use here). 

\begin{figure}[!h]
\vspace{-0.4cm}
\includegraphics[width=0.65\linewidth]{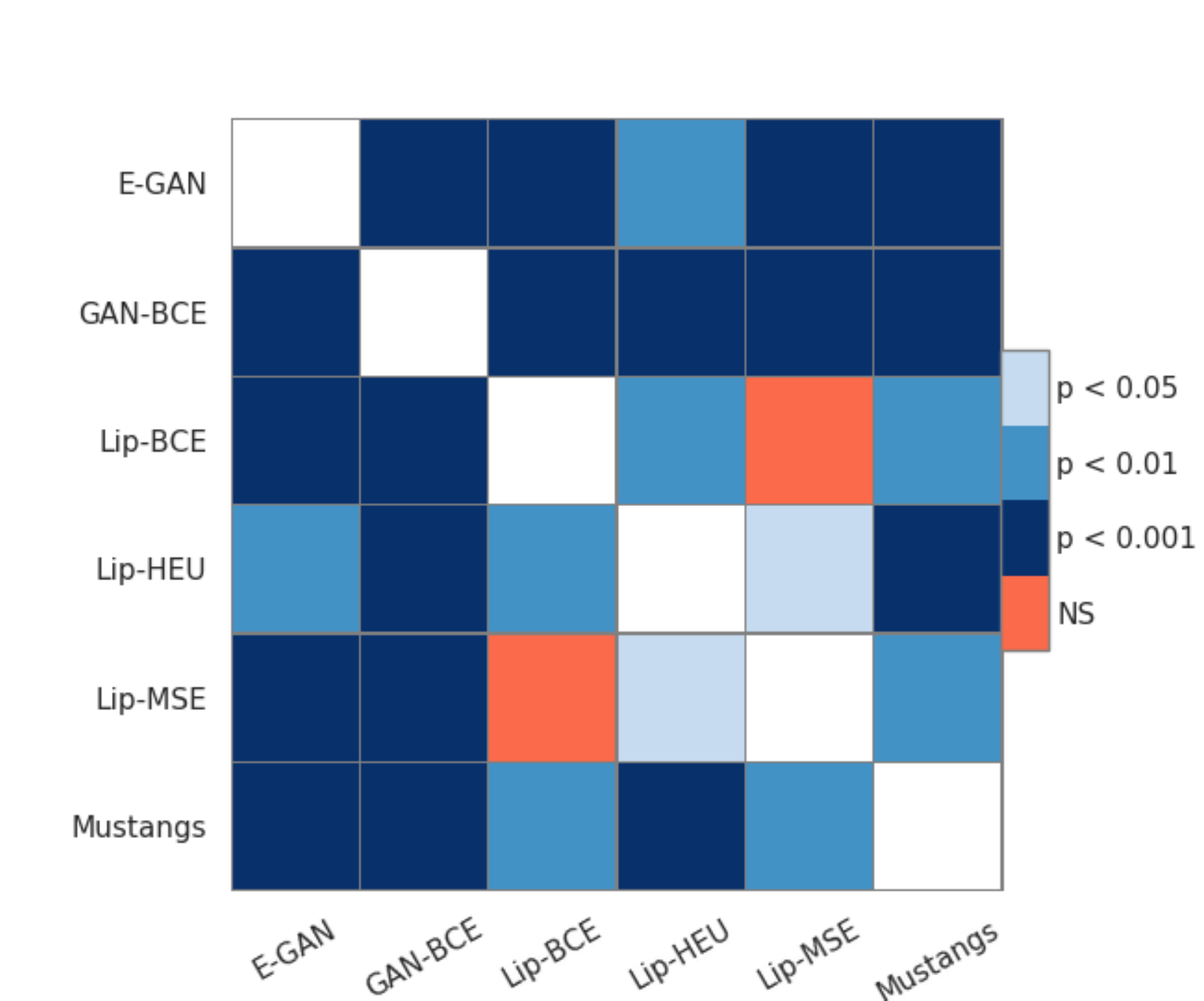}
\vspace{-0.1cm}
\caption{Holm statistical post-hoc analysis on MNIST dataset. It illustrates the $p$-values computed by the statistical tests.}
\vspace{-0.2cm}
\label{fig:fid-holm}
\end{figure}

The results in Table~\ref{tab:fid-results} indicate that spatialy
distributed coevolutionary training is the best choice to train GANs,
even when there is no knowledge about the best loss function to the
problem.  However, the choice of loss function (mutation) may impact
the final results.  In summary, the combination of both mutation and
genome diversity significantly provides the most best result. A
ranksum test with Holm correction confirms that the difference between
\SEGAN and the other methods is significant at confidence levels of
$\alpha$=0.01 and $\alpha$=0.001 (see Figure~\ref{fig:fid-holm}).

\begin{figure}[!h]
\vspace{-0.2cm}
\includegraphics[width=0.96\linewidth]{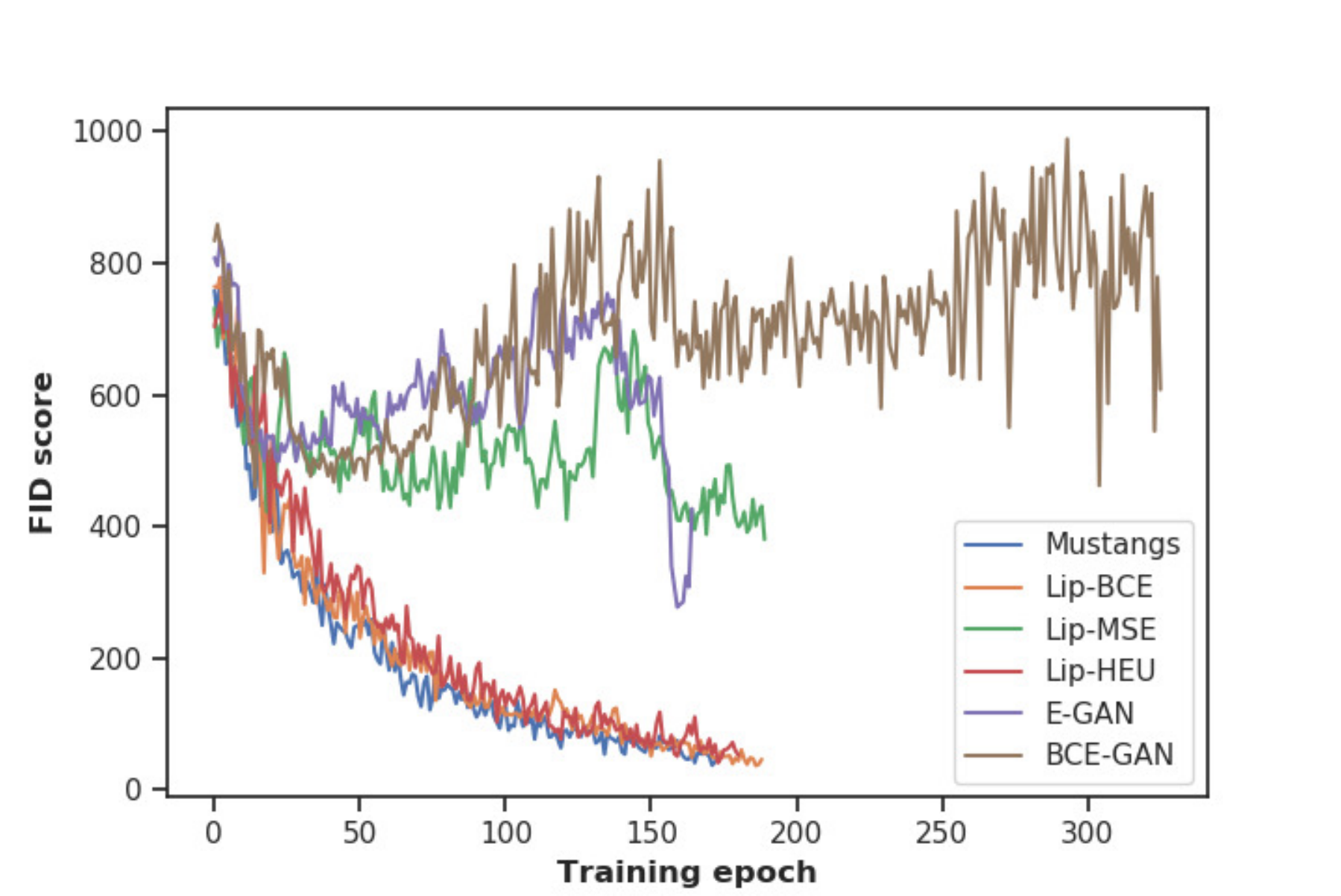}
\vspace{-0.2cm}
\caption{Results on MNIST dataset. FID evolution through the training process during 9h of computational cost.}
\label{fig:fid-evolution}
\vspace{-0.5cm}
\end{figure}

\begin{figure}[!h]
\vspace{-0.2cm}
\includegraphics[width=0.96\linewidth]{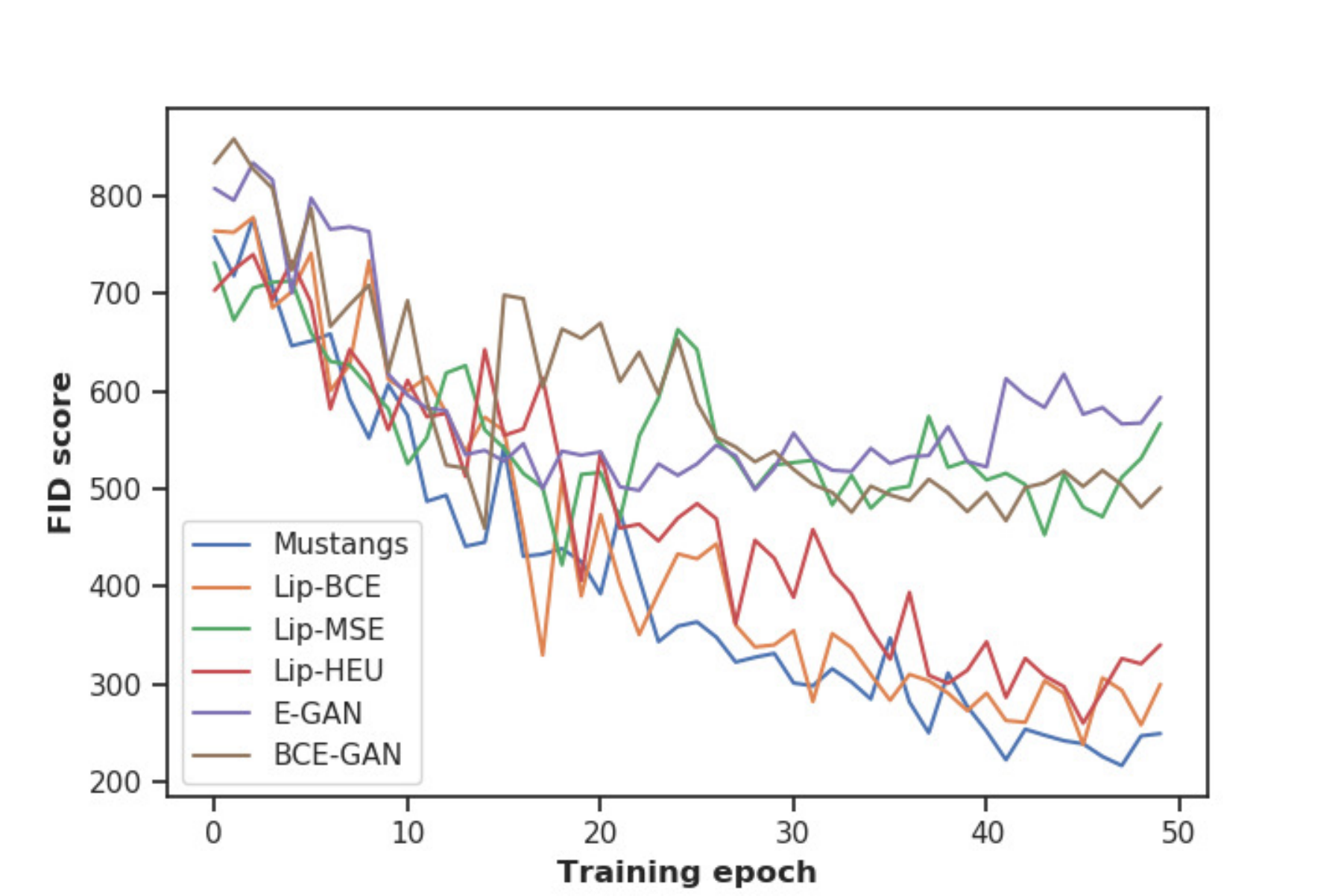}
\vspace{-0.2cm}
\caption{Results on MNIST dataset. FID evolution through the first 50 epochs of the training process.}
\label{fig:fid-evolution-50}
\end{figure}

Next, we evaluate the FID score through out the GAN training process, see
Figure~\ref{fig:fid-evolution} illustrates the FID changes during the
entire training process. In addition, we zoom in on the first 50
training epochs in Figure~\ref{fig:fid-evolution-50}. None of the
evolutionary GAN training methods seem to have converged after 9h of
computation. This implies that longer runs can further improve the
results.

According to Figure~\ref{fig:fid-evolution}, the robustness of the three most competitive methods (\SEGAN, \SCoevGANmm, and \SCoevGANh) indicates that the FID almost behaves like a monotonically decreasing function with small oscilations.
The other three methods have larger osilations and does not seem to have a FID trend that decreases with the same rate. 

Focusing on the two methods that apply the same unique loss function in Equation~(\ref{eq:minmax-mut}), \DCGAN and \SCoevGANmm, we can clearly state the benefits of the distributed spatial evolution. Even the two methods provide comparable FID during the first 30 training epochs, \SCoevGANmm converges faster to better FID values (see Figure~\ref{fig:fid-evolution-50}). This difference is even more noticeable when both algorithms consume the 9h of computational cost (see Figure~\ref{fig:fid-evolution}).

Notice that the spatial coevolutionary methods use FID as the objective function to select the best mixture of generators during the optimization of the GANs. In contrast, \EGAN applies a specific objective function based on the losses~\cite{wang2018evolutionary} and \DCGAN optimizes just one network. 

\subsection{Computational Performance}
\label{sec:perfomrance}

In this section, we analyze the computational performance of the GAN
training methods, all used the same computational budget (9h). 
We start analyzing the number of training epochs. 
As \SEGAN and \Lipizzaner variations apply asynchronous parallelism, the number of training epochs performed by each cell of the grid in the same run varies.
Thus, for these methods, we consider that the number of training epochs of a given run is the mean of the epochs performed by each cell. 

Table~\ref{tab:iter-results} shows the mean, normalized standard
deviation, minimum, and maximum values of number of training epochs.
The number of epochs are normally distributed for all the
algorithms. Please note that, first, all the analyzed methods have
been executed on a cloud architecture, which could generate some
differences in terms of computational efficiency results; and second,
everything is implemented by using the same \texttt{Python} libraries
and versions to limit computational differences between them due to
the technologies used. 
 
\begin{table}[!h]
	\centering
	\small
	\caption{\small Results on MNIST dataset. Mean, normalized standard deviation, minimum, median and maximum of training epochs.}
	\vspace{-0.1cm}
	\label{tab:iter-results}
	\begin{tabular}{lrrrrr}
	    \toprule
		\textbf{Algorithm} & \textbf{Mean} & \textbf{Std\%} & \textbf{Minimum} &\textbf{Median} &  \textbf{Maximum}  \\ \hline
\SEGAN & 179.456 & 2.708\% & 174.778 &  177.944  &  194.000 \\  
\SCoevGANmm & 185.919 & 4.072\% & 171.222 &  189.444  &  194.333 \\ 
\SCoevGANls & 185.963 & 3.768\% & 173.667 &  189.222  &  194.778 \\ 
\SCoevGANh & 188.644 & 4.657\% & 175.667 &  186.667  &  200.556 \\  
\EGAN & 193.167 & 2.992\% & 166.000 &  193.000  &  199.000 \\ 
\DCGAN & \textit{365.267} & 10.570\% & 322.000 &  339.500  &  423.000 \\ 
		\bottomrule
	\end{tabular}
	\vspace{-0.1cm}
\end{table}

\begin{figure}[!h]
\vspace{-0.3cm}
\includegraphics[width=\linewidth]{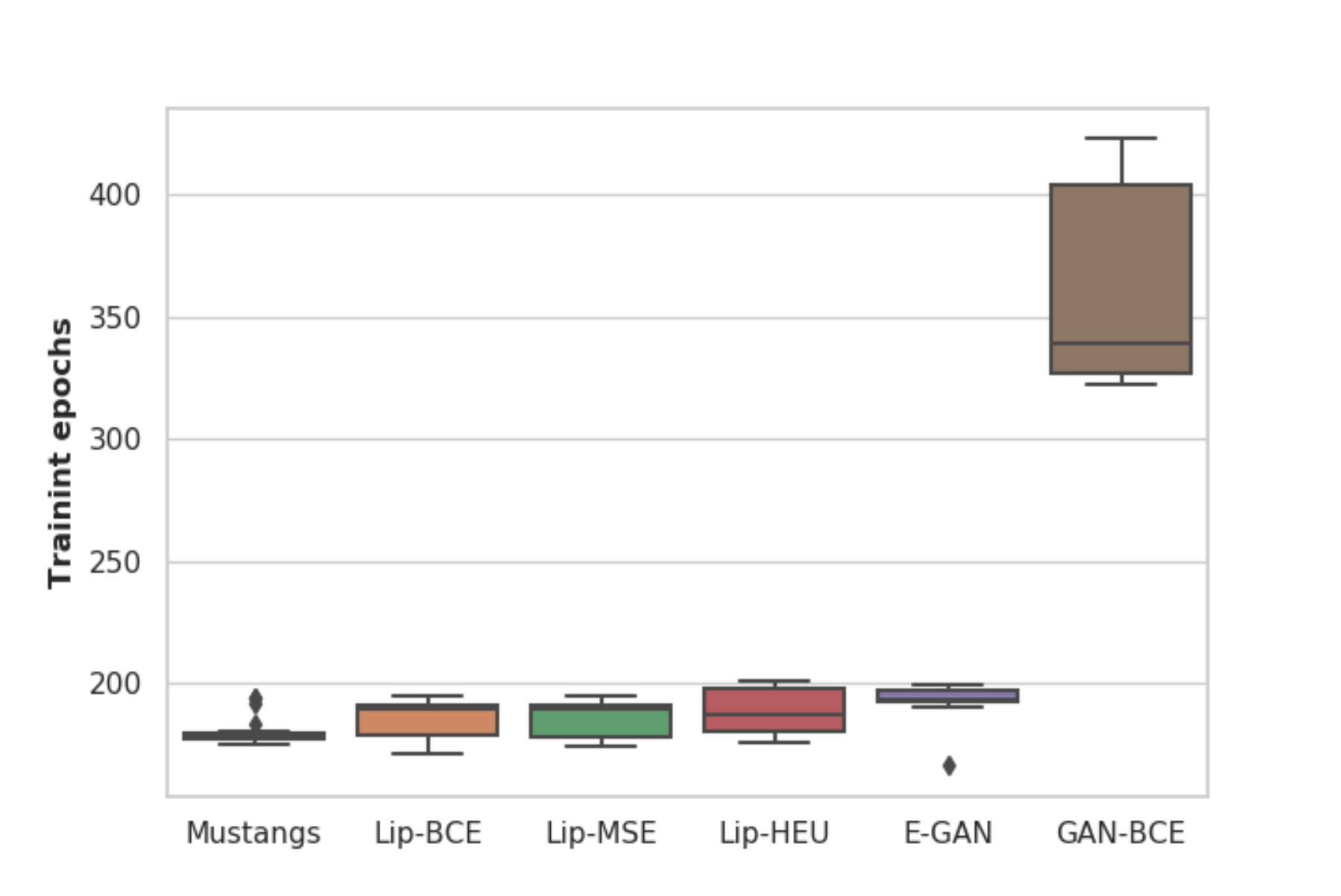}
\vspace{-0.2cm}
\caption{Results on MNIST dataset. Boxplot of the number of training epochs.}
\label{fig:iter-boxplot}
\vspace{-0.1cm}
\end{figure}

The method that is able to train each network for the most epochs is
\DCGAN.  However, this method trains only one network, in
contrast to the other evaluated methods. \DCGAN performs about two
times the number of iterations of \EGAN, which evaluates three
networks.

The spatially distributed coevolutionary algorithms performed
significantly fewer training epochs than \EGAN. However, during each
epoch these methods evaluate 45 GANs, i.e., neighborhood size of 5
$\times$ 9 cells, which is 15 times more networks than \EGAN.

One of the most important features of the spatial coevolutionary
algorithm is that it is executed asynchronously and in parallel for
all the cells~\cite{schmiedlechner2018lipizzaner}.  Thus, there is no
bottleneck for each cells performance since it operates without
waiting for the others. In future work \EGAN could take advantage of
parallelism and optimizing the three discriminators at the same
time. However, it has an important synchronization bottleneck because
they are evaluated over the same discriminator, which is trained and
evaluated sequentially after that operation.

\subsection{Diversity of the Generated Outputs}
\label{sec:diversity}

In this section, we evaluate the diversity of the generated samples by the networks that had the best FID score. 
We report the the total variation distance (TVD) for each algorithm~\citep{li2017distributional} (see Table~\ref{tab:tvd}).

\begin{table}[!h]
\setlength{\tabcolsep}{2pt} 
    \renewcommand{\arraystretch}{1}
	\centering
	\small
	\caption{\small Results on MNIST dataset. TVD results. (Low TVD indicates more diversity)}
	\vspace{-0.1cm}
	\label{tab:tvd}
	\begin{tabular}{lrrrrrr}
	    \toprule
		\textbf{Alg.} & \SEGAN & \SCoevGANmm & \SCoevGANh & \SCoevGANls & \EGAN & \DCGAN \\ \hline
		\textbf{TVD} & 0.180 & 0.171 & \textit{0.115} & 0.365 & 0.534 & 0.516 \\
		\bottomrule
	\end{tabular}
	\vspace{-0.1cm}
\end{table}

\begin{figure*}[!h]
	\begin{tabular}{cccccccc}
	\includegraphics[width=0.12\textwidth, trim={0 63mm 0 0},clip]{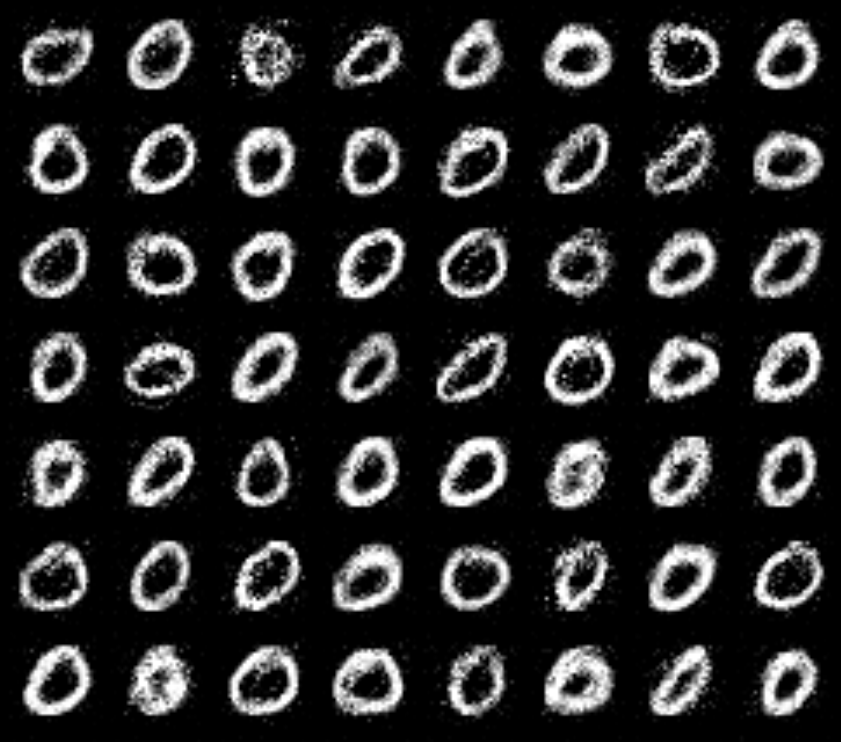} &
		\includegraphics[width=0.12\textwidth, trim={0 63mm 0 0},clip]{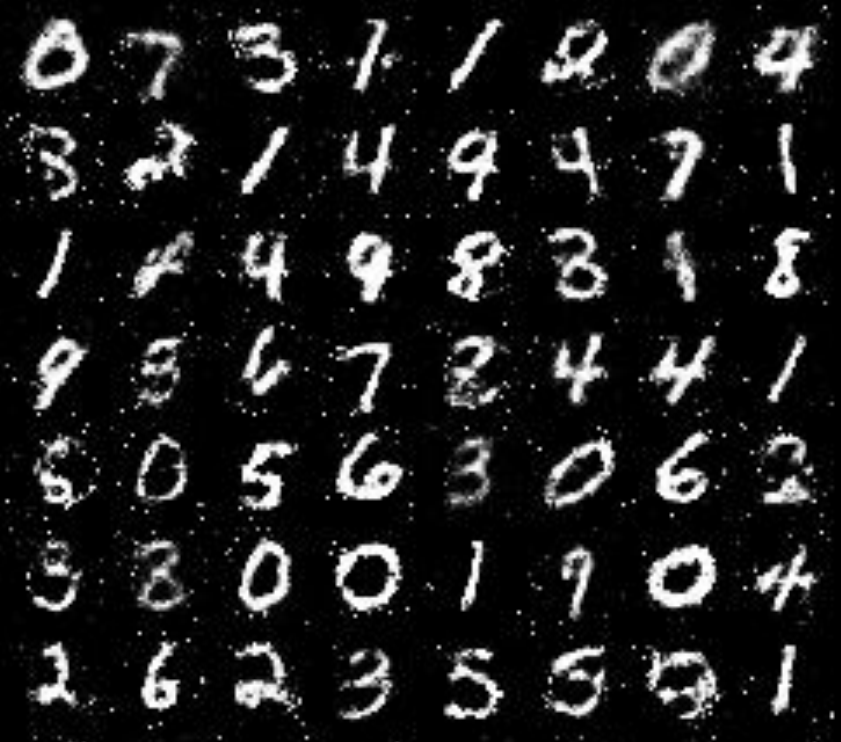} & \includegraphics[width=0.12\textwidth, trim={0 63mm 0 0},clip]{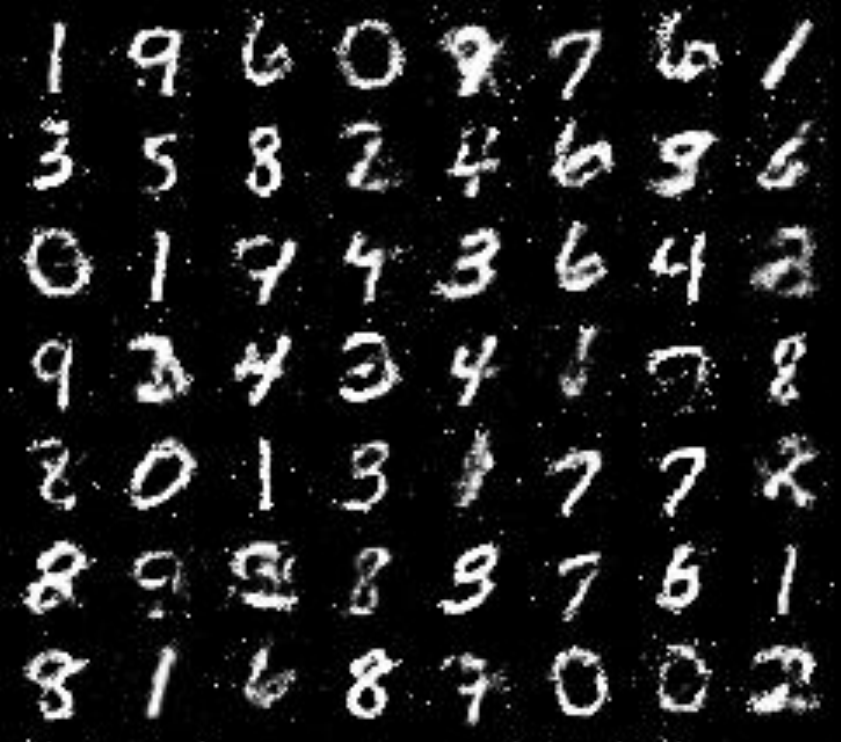} & 
		\includegraphics[width=0.12\textwidth, trim={0 63mm 0 0},clip]{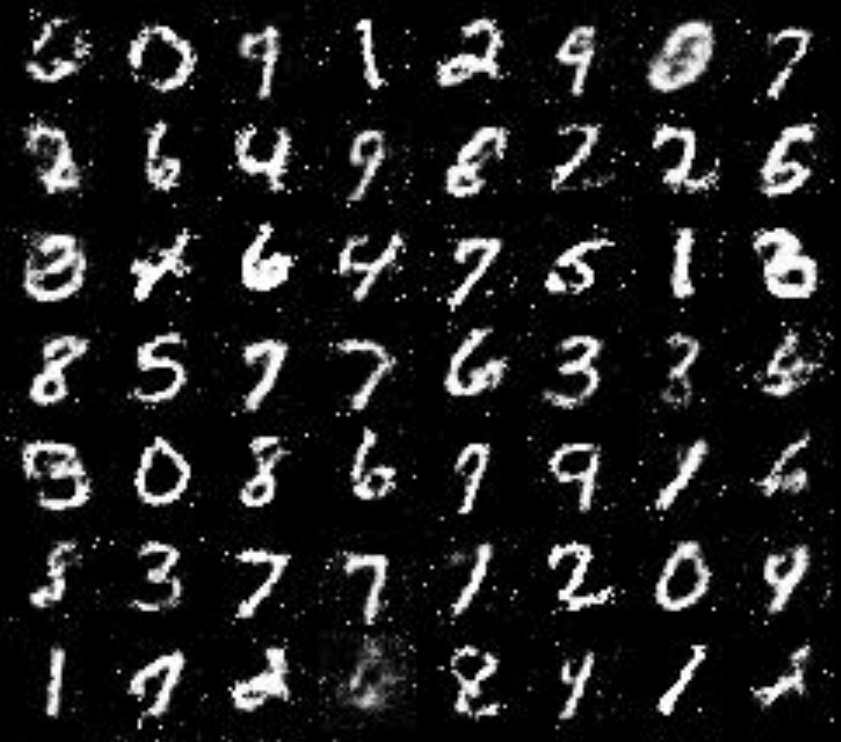} & \includegraphics[width=0.12\textwidth, trim={0 63mm 0 0},clip]{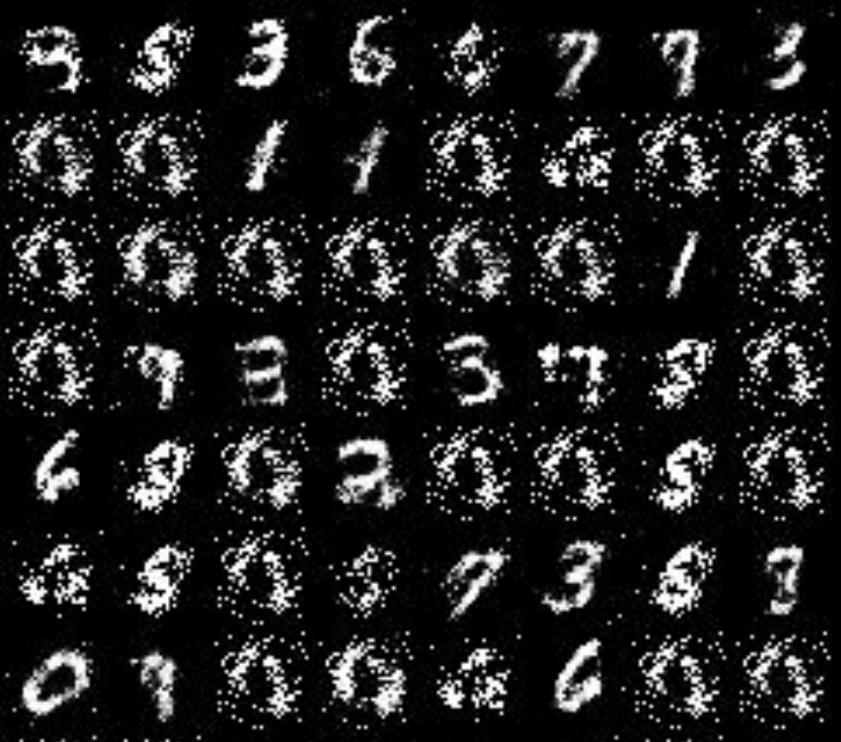} & \includegraphics[width=0.12\textwidth, trim={0 63mm 0 0},clip]{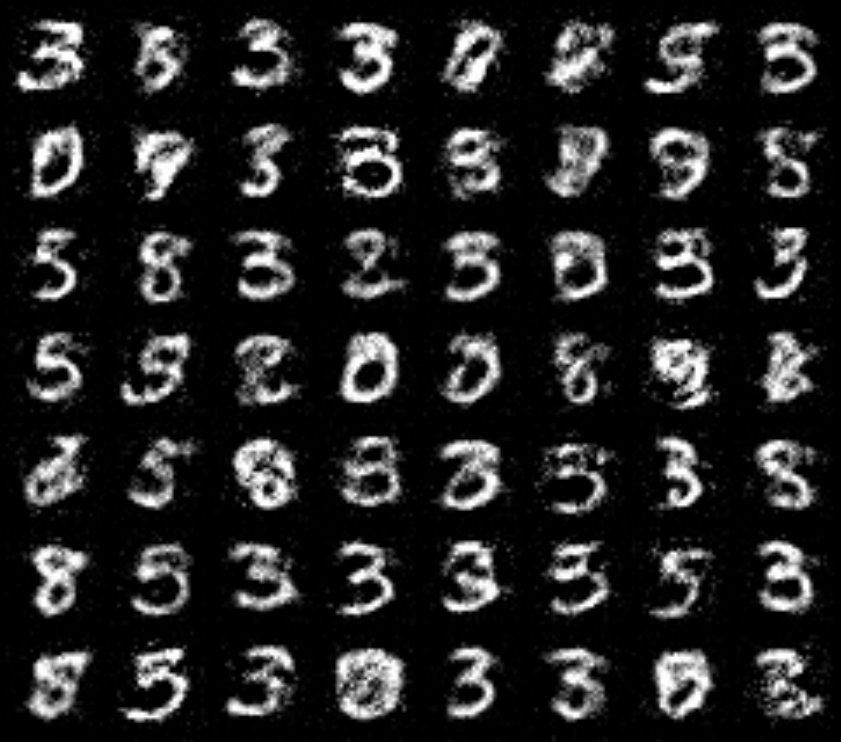} & \includegraphics[width=0.12\textwidth, trim={0 63mm 0 0},clip]{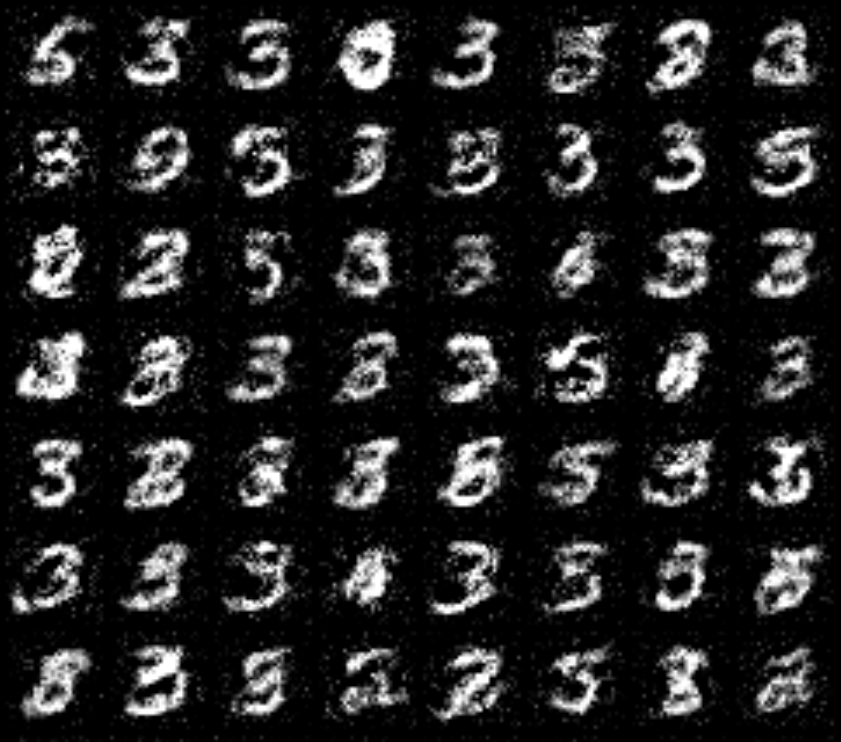} &\\
		(a) Mode collapse & (b) \SEGAN & (c) \SCoevGANmm & (d) \SCoevGANh & (e) \SCoevGANls & (f) \EGAN & (g) \DCGAN
	\end{tabular}	
	\caption{Sequence of samples generated of MNIST dataset. (a) contains for mode collapse, the generator is focused on the character 0. It illustrates samples generated by the best generator (in terms of FID) by \SEGAN (b), \SCoevGANmm (c), \SCoevGANh (d), \SCoevGANls (e), \EGAN (f), and \DCGAN (g). }
	\label{fig:mnist}
\end{figure*}

The methods that provide genome diversity generate more diverse data samples than the other two analyzed methods.
This shows that genomic diversity introduces a capability to avoid mode collapse situations as the one shown in Figure~\ref{fig:mnist}(a). 
The three algorithms with the lowest FID score (\SEGAN, \SCoevGANmm, and \SCoevGANh) also provide the lowest TVD values. The best TVD result is obtained by \SCoevGANh. 

\begin{figure}[!h]
\includegraphics[width=\linewidth]{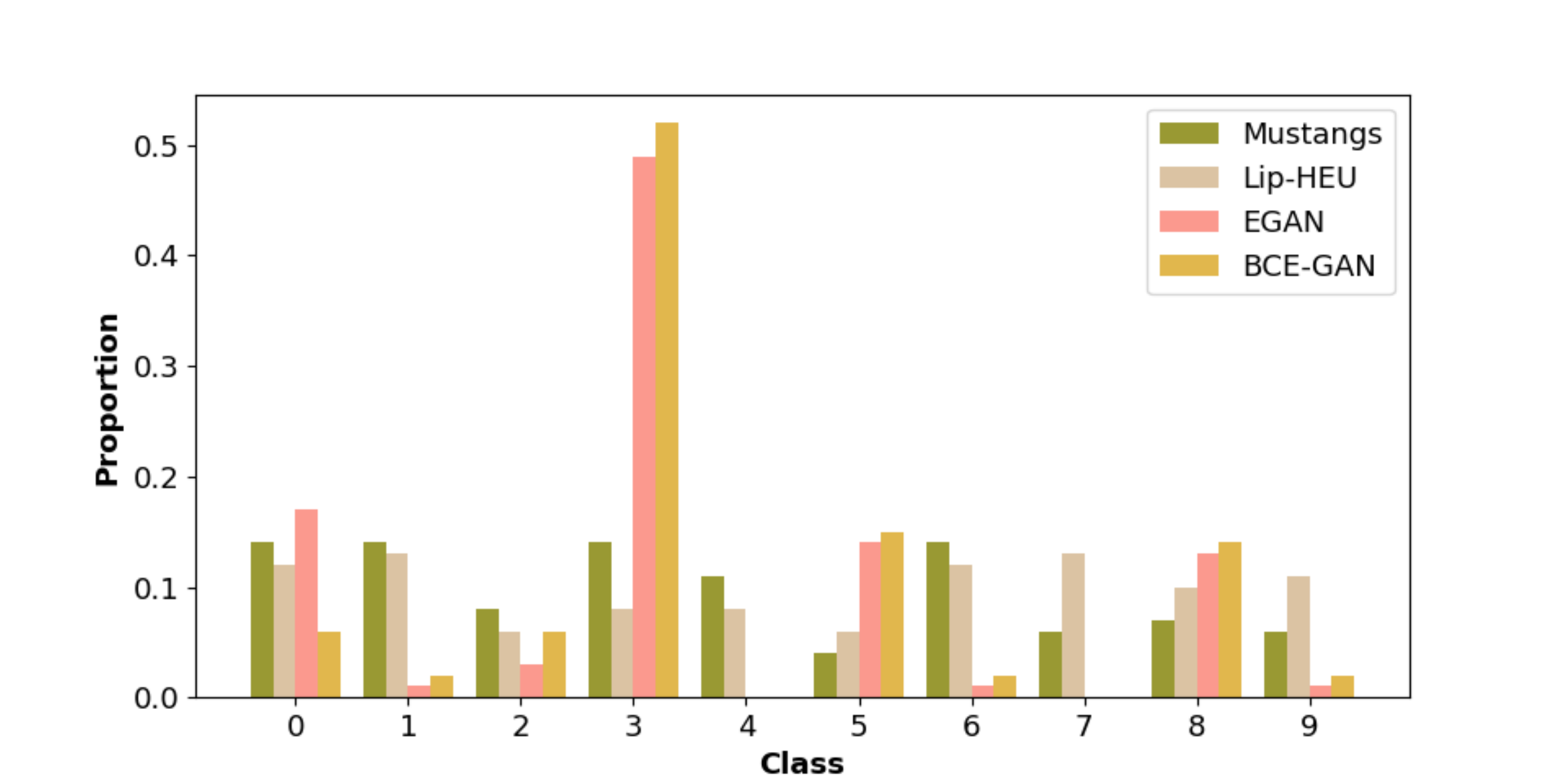}
\caption{Classes distribution of the samples generated of MNIST by \SEGAN, \SCoevGANh, \EGAN, and \DCGAN.}
\label{fig:diversity}
\end{figure}

The distribution of each class of digits for generated images is shown in Figure~\ref{fig:diversity}. 
The diagram supports the TVD results, e.g. \SCoevGANh and \SEGAN produce more diverse set of samples spanning across different classes. We can observe that the two methods that do not apply diversity in terms of genome display a possible mode collapse, since about half of the samples are of class 3 and they not generate samples of class 4 and 7. 
  
Figure~\ref{fig:mnist} illustrates how spatially distributed coevolutionary algorithms are able to produce robust generators that provide with accurate samples across all the classes. 

\subsection{CelebA Experimental Results}

The spatially distributed coevolutionary methods 
are applied to perform the CelebA experiments. 
Table~\ref{tab:fid-results-celeba} summarizes the results over multiple independent runs. 

\begin{table}[!h]
	\centering
	\small
	\caption{\small Results on CelebA dataset. FID results in terms of best mean, normalized standard deviation, median and interquartile range (IQR). (Low FID indicates good performance)}
	\vspace{-0.2cm}
	\label{tab:fid-results-celeba}
	\begin{tabular}{lrrrr}
	    \toprule
		\textbf{Algorithm} & \textbf{Mean} & \textbf{Std\%} & \textbf{Median} & \textbf{IQR}  \\ \hline
\SEGAN & 36.148 & 0.571\% & 36.087 &  0.237 \\ 
\SCoevGANmm & 36.250 & 5.676\% & 36.385 &  2.858 \\ 
\SCoevGANls & 158.740 & 40.226\% & 160.642 &  47.216 \\ 
\SCoevGANh & 37.872 & 5.751\% & 37.681 &  2.455 \\ 
		\bottomrule
	\end{tabular}
	\vspace{-.1cm}
\end{table}

\SEGAN provides the lowest median FID and \SCoevGANls the highest one. 
\SCoevGANmm and \SCoevGANh provide median and average FID scores close to the \SEGAN ones.  
However, \SEGAN is more robust to the varying performance of the methods that apply a unique loss function (see deviations in Table~\ref{tab:fid-results-celeba}). 

The robustness of the training provided by \SEGAN makes it an efficient tool when the computation budget is limited (i.e., performing a limited number of independent runs), since it shows low variability in its competitive results. 

\begin{figure}[!h]
\includegraphics[width=\linewidth]{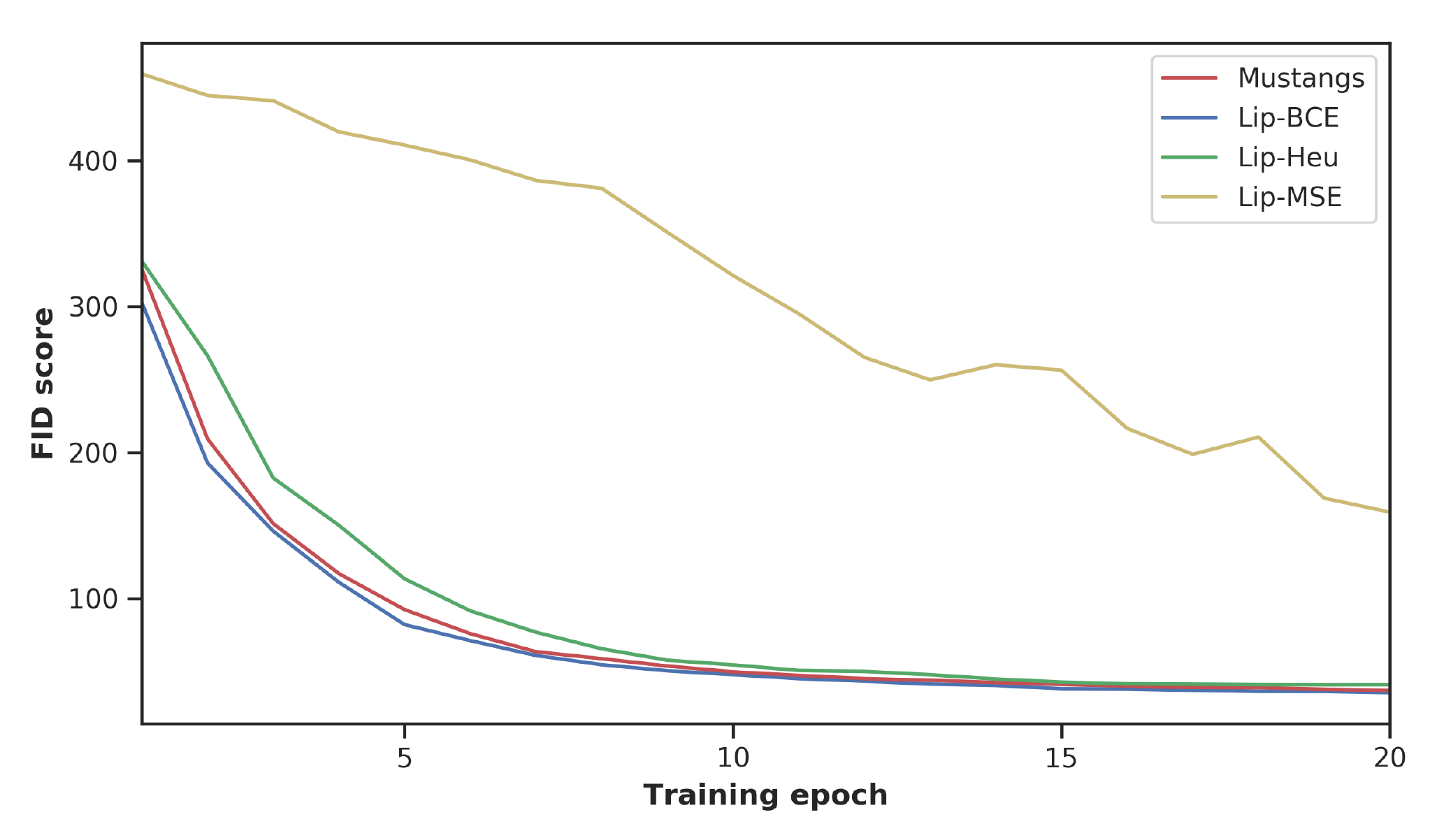}
\caption{Results on CelebA dataset. FID evolution through the 20 epochs of the training process.}
\label{fig:fid-evolution-celeba}
\vspace{-0.2cm}
\end{figure}

Next, we evaluate the FID score through the training process. 
Figure~\ref{fig:fid-evolution-celeba} shows the FID changes during the entire training process. 
All the analyzed methods behave like a monotonically decreasing function. 
However, the FID evolution of \SCoevGANmm presents oscillations. 
\SEGAN, \SCoevGANmm, and \SCoevGANh FIDs show a similar evolution. 

\begin{figure}
	\begin{tabular}{cc}
	\includegraphics[width=0.45\linewidth, trim={0 140mm 0 0},clip]{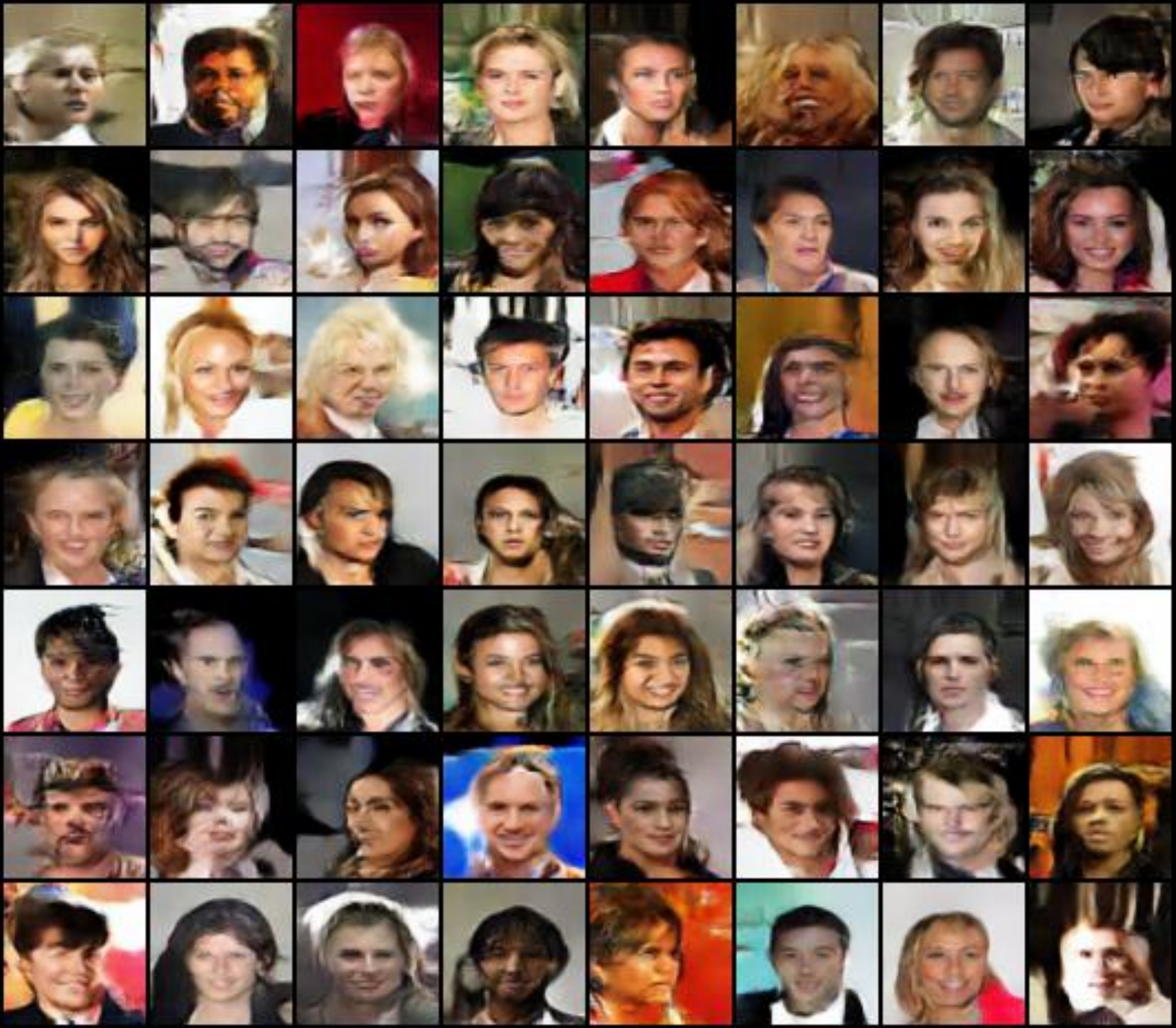} &
		\includegraphics[width=0.45\linewidth, trim={0 140mm 0 0},clip]{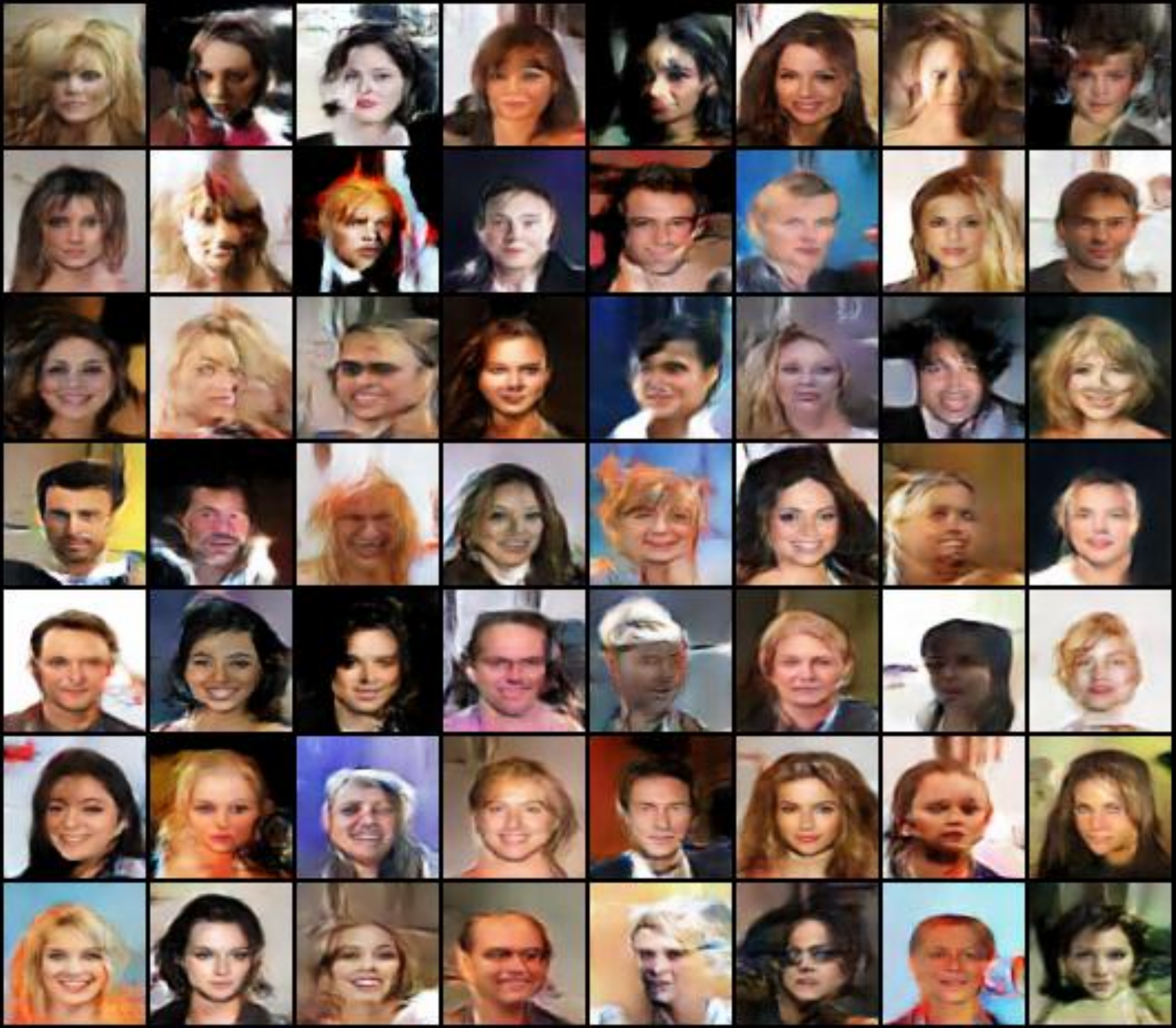} \\
		(a) \SEGAN & (b) \SCoevGANmm 
	\end{tabular}	
	\caption{Sequence of samples generated of CelebA. It illustrates samples generated by the best mixture of generators (in terms of FID) by \SEGAN (a) and \SCoevGANmm (b). }
	\label{fig:celeba-samples}
\end{figure}

Figure~\ref{fig:celeba-samples} illustrates a sequence of samples generated by the best mixture of generators in terms of FID score of the most competitive two training methods, i.e., \SEGAN and \SCoevGANmm. 
As it can be seen in these two sets images generated, both methods present similar capacity of generating human faces. 

%% file: conclusions.tex
\section{Conclusions and Future Work}
\label{sec:conclusions}

We have empirically showed that GAN training can be improved by boosting
diversity. We enhanced an existing spatial evolutionary GAN training
framework that promoted genomic diversity by probabilistically choosing
one of three loss functions. This new method, called \SEGAN was tested
on the MNIST and CelebA data sets showed the best performance and high diversity
in label space, as well as on the TVD measure. The high performance of
\SEGAN is due to its inherent robustness. This allows it to overcome often observed training pathologies, 
e.g.  mode collapse. Furthermore, the \SEGAN method can be executed
asynchronously and the computation is easy to distribute with low
overhead. We extended the \Lipizzaner open source implementation to
demonstrate this.

Future work will include the evaluation of \SEGAN on more data sets and longer training epochs. We can also  include other loss functions. 
In addition, we are exploring the diversity of the networks over their neighborhoods and the whole grid when applying the different methods studied here. 
This study will be a first step in devising an algorithm that self-adapts the probabilities of the different mutations dynamically. 
Finally, other advancements in evolutionary algorithms that can improve the robustness of GAN training, e.g. temporal evolutionary training, need to be considered.

%% file: acks.tex
\section*{Acknowledgments}
\label{sec:acks}
This research was partially funded by European Union’s Horizon 2020 research and innovation programme under the Marie Skłodowska-Curie grant agreement No 799078, and by MINECO and FEDER projects TIN2016-81766-REDT and TIN2017-88213-R. 
The Systems that learn initiative at MIT CSAIL.

%% file: spatial_evolutionary_GAN.bbl

\begin{thebibliography}{26}


\ifx \showCODEN    \undefined \def \showCODEN     #1{\unskip}     \fi
\ifx \showDOI      \undefined \def \showDOI       #1{#1}\fi
\ifx \showISBNx    \undefined \def \showISBNx     #1{\unskip}     \fi
\ifx \showISBNxiii \undefined \def \showISBNxiii  #1{\unskip}     \fi
\ifx \showISSN     \undefined \def \showISSN      #1{\unskip}     \fi
\ifx \showLCCN     \undefined \def \showLCCN      #1{\unskip}     \fi
\ifx \shownote     \undefined \def \shownote      #1{#1}          \fi
\ifx \showarticletitle \undefined \def \showarticletitle #1{#1}   \fi
\ifx \showURL      \undefined \def \showURL       {\relax}        \fi
\providecommand\bibfield[2]{#2}
\providecommand\bibinfo[2]{#2}
\providecommand\natexlab[1]{#1}
\providecommand\showeprint[2][]{arXiv:#2}

\bibitem[\protect\citeauthoryear{Al-Dujaili, Schmiedlechner, Hemberg, and
  O'Reilly}{Al-Dujaili et~al\mbox{.}}{2018}]%
        {schmiedlechner2018towards}
\bibfield{author}{\bibinfo{person}{Abdullah Al-Dujaili}, \bibinfo{person}{Tom
  Schmiedlechner}, \bibinfo{person}{Erik Hemberg}, {and}
  \bibinfo{person}{Una-May O'Reilly}.} \bibinfo{year}{2018}\natexlab{}.
\newblock \showarticletitle{Towards distributed coevolutionary {GANs}}. In
  \bibinfo{booktitle}{{\em AAAI 2018 Fall Symposium}}.
\newblock


\bibitem[\protect\citeauthoryear{Arjovsky and Bottou}{Arjovsky and
  Bottou}{2017}]%
        {arjovsky2017towards}
\bibfield{author}{\bibinfo{person}{Martin Arjovsky} {and}
  \bibinfo{person}{L{\'e}on Bottou}.} \bibinfo{year}{2017}\natexlab{}.
\newblock \showarticletitle{Towards principled methods for training generative
  adversarial networks}.
\newblock \bibinfo{journal}{{\em arXiv preprint arXiv:1701.04862\/}}
  (\bibinfo{year}{2017}).
\newblock


\bibitem[\protect\citeauthoryear{Arjovsky, Chintala, and Bottou}{Arjovsky
  et~al\mbox{.}}{2017}]%
        {arjovsky2017wasserstein}
\bibfield{author}{\bibinfo{person}{Martin Arjovsky}, \bibinfo{person}{Soumith
  Chintala}, {and} \bibinfo{person}{L{\'e}on Bottou}.}
  \bibinfo{year}{2017}\natexlab{}.
\newblock \showarticletitle{Wasserstein GAN}.
\newblock \bibinfo{journal}{{\em arXiv preprint arXiv:1701.07875\/}}
  (\bibinfo{year}{2017}).
\newblock


\bibitem[\protect\citeauthoryear{Arora, Risteski, and Zhang}{Arora
  et~al\mbox{.}}{2018}]%
        {arora2017gans}
\bibfield{author}{\bibinfo{person}{Sanjeev Arora}, \bibinfo{person}{Andrej
  Risteski}, {and} \bibinfo{person}{Yi Zhang}.}
  \bibinfo{year}{2018}\natexlab{}.
\newblock \showarticletitle{Do {GAN}s learn the distribution? Some Theory and
  Empirics}. In \bibinfo{booktitle}{{\em International Conference on Learning
  Representations}}.
\newblock
\showURL{%
\url{https://openreview.net/forum?id=BJehNfW0-}}


\bibitem[\protect\citeauthoryear{Gan, Chen, Wang, Pu, Zhang, Liu, Li, and
  Carin}{Gan et~al\mbox{.}}{2017}]%
        {gan2017triangle}
\bibfield{author}{\bibinfo{person}{Zhe Gan}, \bibinfo{person}{Liqun Chen},
  \bibinfo{person}{Weiyao Wang}, \bibinfo{person}{Yuchen Pu},
  \bibinfo{person}{Yizhe Zhang}, \bibinfo{person}{Hao Liu},
  \bibinfo{person}{Chunyuan Li}, {and} \bibinfo{person}{Lawrence Carin}.}
  \bibinfo{year}{2017}\natexlab{}.
\newblock \showarticletitle{Triangle generative adversarial networks}. In
  \bibinfo{booktitle}{{\em Advances in Neural Information Processing Systems}}.
  \bibinfo{pages}{5247--5256}.
\newblock


\bibitem[\protect\citeauthoryear{Goodfellow, Pouget-Abadie, Mirza, Xu,
  Warde-Farley, Ozair, Courville, and Bengio}{Goodfellow et~al\mbox{.}}{2014}]%
        {goodfellow2014generative}
\bibfield{author}{\bibinfo{person}{Ian Goodfellow}, \bibinfo{person}{Jean
  Pouget-Abadie}, \bibinfo{person}{Mehdi Mirza}, \bibinfo{person}{Bing Xu},
  \bibinfo{person}{David Warde-Farley}, \bibinfo{person}{Sherjil Ozair},
  \bibinfo{person}{Aaron Courville}, {and} \bibinfo{person}{Yoshua Bengio}.}
  \bibinfo{year}{2014}\natexlab{}.
\newblock \showarticletitle{Generative adversarial nets}. In
  \bibinfo{booktitle}{{\em Advances in neural information processing systems}}.
  \bibinfo{pages}{2672--2680}.
\newblock


\bibitem[\protect\citeauthoryear{Heusel, Ramsauer, Unterthiner, Nessler,
  Klambauer, and Hochreiter}{Heusel et~al\mbox{.}}{2017}]%
        {heusel2017gans}
\bibfield{author}{\bibinfo{person}{Martin Heusel}, \bibinfo{person}{Hubert
  Ramsauer}, \bibinfo{person}{Thomas Unterthiner}, \bibinfo{person}{Bernhard
  Nessler}, \bibinfo{person}{G{\"u}nter Klambauer}, {and} \bibinfo{person}{Sepp
  Hochreiter}.} \bibinfo{year}{2017}\natexlab{}.
\newblock \showarticletitle{{GAN}s trained by a two time-scale update rule
  converge to a nash equilibrium}.
\newblock \bibinfo{journal}{{\em arXiv preprint arXiv:1706.08500\/}}
  \bibinfo{volume}{12}, \bibinfo{number}{1} (\bibinfo{year}{2017}).
\newblock


\bibitem[\protect\citeauthoryear{Jiwoong~Im, Ma, Dongjoo~Kim, and
  Taylor}{Jiwoong~Im et~al\mbox{.}}{2016}]%
        {jiwoong2016generative}
\bibfield{author}{\bibinfo{person}{Daniel Jiwoong~Im}, \bibinfo{person}{He Ma},
  \bibinfo{person}{Chris Dongjoo~Kim}, {and} \bibinfo{person}{Graham Taylor}.}
  \bibinfo{year}{2016}\natexlab{}.
\newblock \showarticletitle{Generative Adversarial Parallelization}.
\newblock \bibinfo{journal}{{\em arXiv preprint arXiv:1612.04021\/}}
  (\bibinfo{year}{2016}).
\newblock


\bibitem[\protect\citeauthoryear{Li, Alvarez-Melis, Xu, Jegelka, and Sra}{Li
  et~al\mbox{.}}{2017a}]%
        {li2017distributional}
\bibfield{author}{\bibinfo{person}{Chengtao Li}, \bibinfo{person}{David
  Alvarez-Melis}, \bibinfo{person}{Keyulu Xu}, \bibinfo{person}{Stefanie
  Jegelka}, {and} \bibinfo{person}{Suvrit Sra}.}
  \bibinfo{year}{2017}\natexlab{a}.
\newblock \showarticletitle{Distributional Adversarial Networks}.
\newblock \bibinfo{journal}{{\em arXiv preprint arXiv:1706.09549\/}}
  (\bibinfo{year}{2017}).
\newblock


\bibitem[\protect\citeauthoryear{Li, Madry, Peebles, and Schmidt}{Li
  et~al\mbox{.}}{2017b}]%
        {li2017towards}
\bibfield{author}{\bibinfo{person}{Jerry Li}, \bibinfo{person}{Aleksander
  Madry}, \bibinfo{person}{John Peebles}, {and} \bibinfo{person}{Ludwig
  Schmidt}.} \bibinfo{year}{2017}\natexlab{b}.
\newblock \showarticletitle{Towards Understanding the Dynamics of Generative
  Adversarial Networks}.
\newblock \bibinfo{journal}{{\em arXiv preprint arXiv:1706.09884\/}}
  (\bibinfo{year}{2017}).
\newblock


\bibitem[\protect\citeauthoryear{Liang, Lee, Dai, and Xing}{Liang
  et~al\mbox{.}}{2017}]%
        {liang2017dual}
\bibfield{author}{\bibinfo{person}{Xiaodan Liang}, \bibinfo{person}{Lisa Lee},
  \bibinfo{person}{Wei Dai}, {and} \bibinfo{person}{Eric~P Xing}.}
  \bibinfo{year}{2017}\natexlab{}.
\newblock \showarticletitle{Dual motion GAN for future-flow embedded video
  prediction}. In \bibinfo{booktitle}{{\em IEEE International Conference on
  Computer Vision (ICCV)}}, Vol.~\bibinfo{volume}{1}.
\newblock


\bibitem[\protect\citeauthoryear{Mao, Li, Xie, Lau, Wang, and Paul~Smolley}{Mao
  et~al\mbox{.}}{2017}]%
        {mao2017least}
\bibfield{author}{\bibinfo{person}{Xudong Mao}, \bibinfo{person}{Qing Li},
  \bibinfo{person}{Haoran Xie}, \bibinfo{person}{Raymond~YK Lau},
  \bibinfo{person}{Zhen Wang}, {and} \bibinfo{person}{Stephen Paul~Smolley}.}
  \bibinfo{year}{2017}\natexlab{}.
\newblock \showarticletitle{Least squares generative adversarial networks}. In
  \bibinfo{booktitle}{{\em Proceedings of the IEEE International Conference on
  Computer Vision}}. \bibinfo{pages}{2794--2802}.
\newblock


\bibitem[\protect\citeauthoryear{Mitchell, Thomure, and Williams}{Mitchell
  et~al\mbox{.}}{2006}]%
        {Mitchell06therole}
\bibfield{author}{\bibinfo{person}{Melanie Mitchell},
  \bibinfo{person}{Michael~D. Thomure}, {and} \bibinfo{person}{Nathan~L.
  Williams}.} \bibinfo{year}{2006}\natexlab{}.
\newblock \showarticletitle{The role of space in the success of coevolutionary
  learning}. In \bibinfo{booktitle}{{\em Artificial Life X: Proceedings of the
  Tenth International Conference on the Simulation and Synthesis of Living
  Systems}}. \bibinfo{publisher}{MIT Press}, \bibinfo{pages}{118--124}.
\newblock


\bibitem[\protect\citeauthoryear{Neyshabur, Bhojanapalli, and
  Chakrabarti}{Neyshabur et~al\mbox{.}}{2017}]%
        {neyshabur2017stabilizing}
\bibfield{author}{\bibinfo{person}{Behnam Neyshabur}, \bibinfo{person}{Srinadh
  Bhojanapalli}, {and} \bibinfo{person}{Ayan Chakrabarti}.}
  \bibinfo{year}{2017}\natexlab{}.
\newblock \showarticletitle{Stabilizing GAN training with multiple random
  projections}.
\newblock \bibinfo{journal}{{\em arXiv preprint arXiv:1705.07831\/}}
  (\bibinfo{year}{2017}).
\newblock


\bibitem[\protect\citeauthoryear{Nguyen, Le, Vu, and Phung}{Nguyen
  et~al\mbox{.}}{2017}]%
        {nguyen2017dual}
\bibfield{author}{\bibinfo{person}{Tu Nguyen}, \bibinfo{person}{Trung Le},
  \bibinfo{person}{Hung Vu}, {and} \bibinfo{person}{Dinh Phung}.}
  \bibinfo{year}{2017}\natexlab{}.
\newblock \showarticletitle{Dual discriminator generative adversarial nets}. In
  \bibinfo{booktitle}{{\em Advances in Neural Information Processing Systems}}.
  \bibinfo{pages}{2670--2680}.
\newblock


\bibitem[\protect\citeauthoryear{Nielsen, Dorronsoro, Danoy, and
  Bouvry}{Nielsen et~al\mbox{.}}{2012}]%
        {nielsen2012novel}
\bibfield{author}{\bibinfo{person}{Sune~S Nielsen},
  \bibinfo{person}{Bernab{\'e} Dorronsoro}, \bibinfo{person}{Gr{\'e}goire
  Danoy}, {and} \bibinfo{person}{Pascal Bouvry}.}
  \bibinfo{year}{2012}\natexlab{}.
\newblock \showarticletitle{Novel efficient asynchronous cooperative
  co-evolutionary multi-objective algorithms}. In \bibinfo{booktitle}{{\em
  Evolutionary Computation (CEC), 2012 IEEE Congress on}}. IEEE,
  \bibinfo{pages}{1--7}.
\newblock


\bibitem[\protect\citeauthoryear{Popovici, Bucci, Wiegand, and
  De~Jong}{Popovici et~al\mbox{.}}{2012}]%
        {popovici2012coevolutionary}
\bibfield{author}{\bibinfo{person}{Elena Popovici}, \bibinfo{person}{Anthony
  Bucci}, \bibinfo{person}{R~Paul Wiegand}, {and} \bibinfo{person}{Edwin~D
  De~Jong}.} \bibinfo{year}{2012}\natexlab{}.
\newblock \showarticletitle{Coevolutionary principles}.
\newblock In \bibinfo{booktitle}{{\em Handbook of natural computing}}.
  \bibinfo{publisher}{Springer}, \bibinfo{pages}{987--1033}.
\newblock


\bibitem[\protect\citeauthoryear{Radford, Metz, and Chintala}{Radford
  et~al\mbox{.}}{2015}]%
        {Radford2015unsupervised}
\bibfield{author}{\bibinfo{person}{Alec Radford}, \bibinfo{person}{Luke Metz},
  {and} \bibinfo{person}{Soumith Chintala}.} \bibinfo{year}{2015}\natexlab{}.
\newblock \showarticletitle{Unsupervised Representation Learning with Deep
  Convolutional Generative Adversarial Networks}.
\newblock \bibinfo{journal}{{\em arXiv preprint arXiv:1511.06434\/}}
  (\bibinfo{year}{2015}).
\newblock


\bibitem[\protect\citeauthoryear{Reed, Akata, Mohan, Tenka, Schiele, and
  Lee}{Reed et~al\mbox{.}}{2016}]%
        {reed2016learning}
\bibfield{author}{\bibinfo{person}{Scott~E Reed}, \bibinfo{person}{Zeynep
  Akata}, \bibinfo{person}{Santosh Mohan}, \bibinfo{person}{Samuel Tenka},
  \bibinfo{person}{Bernt Schiele}, {and} \bibinfo{person}{Honglak Lee}.}
  \bibinfo{year}{2016}\natexlab{}.
\newblock \showarticletitle{Learning what and where to draw}. In
  \bibinfo{booktitle}{{\em Advances in Neural Information Processing Systems}}.
  \bibinfo{pages}{217--225}.
\newblock


\bibitem[\protect\citeauthoryear{Schmiedlechner, Yong, Al-Dujaili, Hemberg, and
  O'Reilly}{Schmiedlechner et~al\mbox{.}}{2018}]%
        {schmiedlechner2018lipizzaner}
\bibfield{author}{\bibinfo{person}{Tom Schmiedlechner},
  \bibinfo{person}{Ignavier Ng~Zhi Yong}, \bibinfo{person}{Abdullah
  Al-Dujaili}, \bibinfo{person}{Erik Hemberg}, {and} \bibinfo{person}{Una-May
  O'Reilly}.} \bibinfo{year}{2018}\natexlab{}.
\newblock \showarticletitle{{Lipizzaner: A System That Scales Robust Generative
  Adversarial Network Training}}. In \bibinfo{booktitle}{{\em the 32nd
  Conference on Neural Information Processing Systems (NeurIPS 2018) Workshop
  on Systems for ML and Open Source Software}}.
\newblock


\bibitem[\protect\citeauthoryear{Tolstikhin, Gelly, Bousquet, Simon-Gabriel,
  and Sch{\"o}lkopf}{Tolstikhin et~al\mbox{.}}{2017}]%
        {tolstikhin2017adagan}
\bibfield{author}{\bibinfo{person}{Ilya~O Tolstikhin}, \bibinfo{person}{Sylvain
  Gelly}, \bibinfo{person}{Olivier Bousquet}, \bibinfo{person}{Carl-Johann
  Simon-Gabriel}, {and} \bibinfo{person}{Bernhard Sch{\"o}lkopf}.}
  \bibinfo{year}{2017}\natexlab{}.
\newblock \showarticletitle{Adagan: Boosting generative models}. In
  \bibinfo{booktitle}{{\em Advances in Neural Information Processing Systems}}.
  \bibinfo{pages}{5430--5439}.
\newblock


\bibitem[\protect\citeauthoryear{Wang, Xu, Yao, and Tao}{Wang
  et~al\mbox{.}}{2018}]%
        {wang2018evolutionary}
\bibfield{author}{\bibinfo{person}{Chaoyue Wang}, \bibinfo{person}{Chang Xu},
  \bibinfo{person}{Xin Yao}, {and} \bibinfo{person}{Dacheng Tao}.}
  \bibinfo{year}{2018}\natexlab{}.
\newblock \showarticletitle{Evolutionary Generative Adversarial Networks}.
\newblock \bibinfo{journal}{{\em arXiv preprint arXiv:1803.00657\/}}
  (\bibinfo{year}{2018}).
\newblock


\bibitem[\protect\citeauthoryear{Wang, Zhang, and van~de Weijer}{Wang
  et~al\mbox{.}}{2016}]%
        {wang2016ensembles}
\bibfield{author}{\bibinfo{person}{Yaxing Wang}, \bibinfo{person}{Lichao
  Zhang}, {and} \bibinfo{person}{Joost van~de Weijer}.}
  \bibinfo{year}{2016}\natexlab{}.
\newblock \showarticletitle{Ensembles of generative adversarial networks}.
\newblock \bibinfo{journal}{{\em arXiv preprint arXiv:1612.00991\/}}
  (\bibinfo{year}{2016}).
\newblock


\bibitem[\protect\citeauthoryear{Williams and Mitchell}{Williams and
  Mitchell}{2005}]%
        {Williams2005}
\bibfield{author}{\bibinfo{person}{Nathan Williams} {and}
  \bibinfo{person}{Melanie Mitchell}.} \bibinfo{year}{2005}\natexlab{}.
\newblock \showarticletitle{Investigating the Success of Spatial Coevolution}.
  In \bibinfo{booktitle}{{\em Proceedings of the 7th Annual Conference on
  Genetic and Evolutionary Computation}} {\em (\bibinfo{series}{GECCO '05})}.
  \bibinfo{publisher}{ACM}, \bibinfo{address}{New York, NY, USA},
  \bibinfo{pages}{523--530}.
\newblock
\showISBNx{1-59593-010-8}
\showDOI{%
\url{https://doi.org/10.1145/1068009.1068096}}


\bibitem[\protect\citeauthoryear{Yeh, Chen, Yian~Lim, Schwing,
  Hasegawa-Johnson, and Do}{Yeh et~al\mbox{.}}{2017}]%
        {yeh2017semantic}
\bibfield{author}{\bibinfo{person}{Raymond~A Yeh}, \bibinfo{person}{Chen Chen},
  \bibinfo{person}{Teck Yian~Lim}, \bibinfo{person}{Alexander~G Schwing},
  \bibinfo{person}{Mark Hasegawa-Johnson}, {and} \bibinfo{person}{Minh~N Do}.}
  \bibinfo{year}{2017}\natexlab{}.
\newblock \showarticletitle{Semantic image inpainting with deep generative
  models}. In \bibinfo{booktitle}{{\em Proceedings of the IEEE Conference on
  Computer Vision and Pattern Recognition}}. \bibinfo{pages}{5485--5493}.
\newblock


\bibitem[\protect\citeauthoryear{Zhao, Mathieu, and LeCun}{Zhao
  et~al\mbox{.}}{2016}]%
        {zhao2016energy}
\bibfield{author}{\bibinfo{person}{Junbo Zhao}, \bibinfo{person}{Michael
  Mathieu}, {and} \bibinfo{person}{Yann LeCun}.}
  \bibinfo{year}{2016}\natexlab{}.
\newblock \showarticletitle{Energy-based generative adversarial network}.
\newblock \bibinfo{journal}{{\em arXiv preprint arXiv:1609.03126\/}}
  (\bibinfo{year}{2016}).
\newblock


\end{thebibliography}
